\documentclass[10pt,journal,compsoc]{IEEEtran}
\ifCLASSOPTIONcompsoc
  \usepackage[nocompress]{cite}
\else
  \usepackage{cite}
\fi
\usepackage{hyperref} 
\hypersetup{
    colorlinks=true,
    linkcolor=red,
}
\usepackage{amssymb}
\usepackage{bbding}

\usepackage{booktabs}

\ifCLASSINFOpdf
   \usepackage[pdftex]{graphicx}
\else
\fi

\usepackage{stfloats}
\usepackage{color}
\usepackage[dvipsnames]{xcolor}

\usepackage{multirow}

\usepackage{caption}
\usepackage{subcaption}
\usepackage{amsmath}

\usepackage{amssymb}
\usepackage{pifont}
\newcommand{\cmark}{\ding{51}}%
\newcommand{\xmark}{\ding{55}}%

\begin{document}
%
\title{
Vision-Language Models for Vision Tasks: \\ A Survey
}

\author{Jingyi~Zhang$^\dagger$,
        Jiaxing~Huang$^\dagger$,
        Sheng~Jin
        and~Shijian~Lu$^*$
        
\IEEEcompsocitemizethanks{\IEEEcompsocthanksitem All authors are with the School of Computer Science and Engineering, Nanyang Technological University, Singapore.\protect
\IEEEcompsocthanksitem $\dagger$ denotes equal contribution; $*$ denotes corresponding author.}
}

%
%

\markboth{Journal of \LaTeX\ Class Files,
March~2023}%
{Shell \MakeLowercase{\textit{et al.}}: Bare Demo of IEEEtran.cls for Computer Society Journals}
%



\IEEEtitleabstractindextext{%
\begin{abstract} 
Most visual recognition studies rely heavily on crowd-labelled data in deep neural networks (DNNs) training, and they usually train a DNN for each single visual recognition task, leading to a laborious and time-consuming visual recognition paradigm.
To address the two challenges, Vision-Language Models (VLMs) have been intensively investigated recently, which learns rich vision-language correlation from web-scale image-text pairs that are almost infinitely available on the Internet and enables zero-shot predictions on various visual recognition tasks with a single VLM.
This paper provides a systematic review of visual language models for various visual recognition tasks, including: (1) the background that introduces the development of visual recognition paradigms; (2) the foundations of VLM that summarize the widely-adopted network architectures, pre-training objectives, and downstream tasks; (3) the widely-adopted datasets in VLM pre-training and evaluations; (4) the review and categorization of existing VLM pre-training methods, VLM transfer learning methods, and VLM knowledge distillation methods; (5) the benchmarking, analysis and discussion of the reviewed methods; (6) several research challenges and potential research directions that could be pursued in the future VLM studies for visual recognition. A project associated with this survey has been created at \url{https://github.com/jingyi0000/VLM_survey}.
\end{abstract}

}

\maketitle

\IEEEdisplaynontitleabstractindextext

%
\IEEEpeerreviewmaketitle

\IEEEraisesectionheading{\section{Introduction}\label{sec:introduction}}

Visual recognition ($e.g.$, image classification, object detection and semantic segmentation) is a long-standing challenge in computer vision research, and it is also the cornerstone of a myriad of computer vision applications in autonomous driving~\cite{geiger2012we}, remote sensing~\cite{cheng2020remote}, robotics~\cite{pierson2017deep}, etc. With the advent of deep learning~\cite{krizhevsky2017imagenet, simonyan2014very,he2016deep}, visual recognition research has achieved great success by leveraging end-to-end trainable deep neural networks (DNNs). However, the shift from \textit{Traditional Machine Learning}~\cite{peterson2009k,cortes1995support,mathur2008multiclass} toward deep learning comes with two new grand challenges, namely, the slow convergence of DNN training under the classical setup of \textit{Deep Learning from Scratch}~\cite{krizhevsky2017imagenet,simonyan2014very,he2016deep} and the laborious collection of large-scale, task-specific, and crowd-labelled data~\cite{radford2021learning} in DNN training.

Recently, a new learning paradigm \textit{Pre-training, Fine-tuning and Prediction} has demonstrated great effectiveness in a wide range of visual recognition tasks~\cite{girshick2015fast,he2020momentum,chen2020simple}. Under this new paradigm, a DNN model is first pre-trained with certain off-the-shelf large-scale training data, being annotated or unannotated, and the pre-trained model is then fine-tuned with 
task-specific annotated training data as illustrated in Figs.~\ref{fig.intro} (a) and (b). With comprehensive knowledge learned in the pre-trained models, this learning paradigm can accelerate network convergence and train well-performing models for various downstream tasks.

\begin{figure}[ht]
    \centering
    \includegraphics[width=0.42\textwidth]{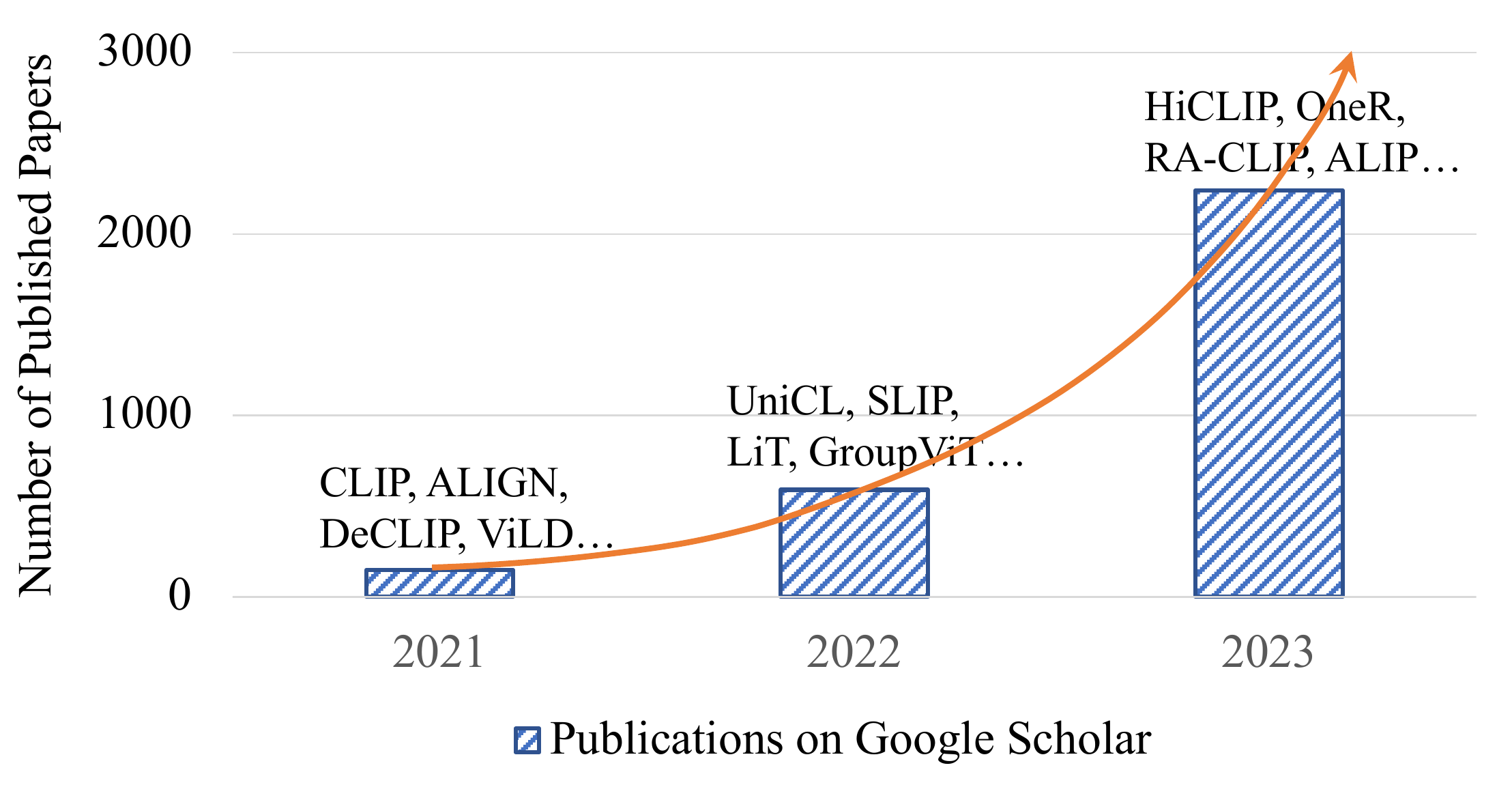}
    \caption{Number of publications on visual recognition VLMs (from Google Scholar). The publications grow exponentially since the pioneer study CLIP~\cite{radford2021learning} in 2021. 
    }
    \label{fig:trend}
\end{figure}

Nevertheless, the \textit{Pre-training, Fine-tuning and Prediction} paradigm still requires an additional stage of task-specific fine-tuning with labelled training data from each downstream task.
Inspired by the advances in natural language processing~\cite{devlin2018bert,radford2018improving,radford2019language}, a new deep learning paradigm named \textit{Vision-Language Model Pre-training and Zero-shot Prediction} has attracted increasing attention recently~\cite{radford2021learning,jia2021scaling,yao2021filip}. 
In this paradigm, a vision-language model (VLM) is pre-trained with large-scale image-text pairs that are almost infinitely available on the internet, and the pre-trained VLM can be directly applied to downstream visual recognition tasks without fine-tuning as illustrated in Fig.~\ref{fig.intro} (c). The VLM pre-training is usually guided by certain vision-language objectives~\cite{radford2021learning,yao2021filip,yu2022coca} that enable to learn image-text correspondences from the large-scale image-text pairs~\cite{schuhmann2022laion,schuhmann2021laion}, $e.g.$, CLIP~\cite{radford2021learning} employs an image-text contrastive objective and learns by pulling the paired images and texts close and pushing others faraway in the embedding space. In this way, the pre-trained VLMs capture rich vision-language correspondence knowledge and can perform zero-shot predictions by matching the embeddings of any given images and texts. 
This new learning paradigm enables effective usage of web data and allows zero-shot predictions without task-specific fine-tuning, which 
is simple to implement yet performs incredibly well, $e.g.$, the pre-trained CLIP has achieved superior zero-shot performance on 36 visual recognition tasks ranging from classic image classification~\cite{bossard2014food, krizhevsky2009learning, xiao2010sun,krause2013collecting,parkhi2012cats} to human action and optical character recognition~\cite{kiela2020hateful,radford2021learning, miech2020rareact, soomro2012ucf101,carreira2019short}.

\begin{figure*}[t]
    \centering
    \includegraphics[width=0.9\textwidth]{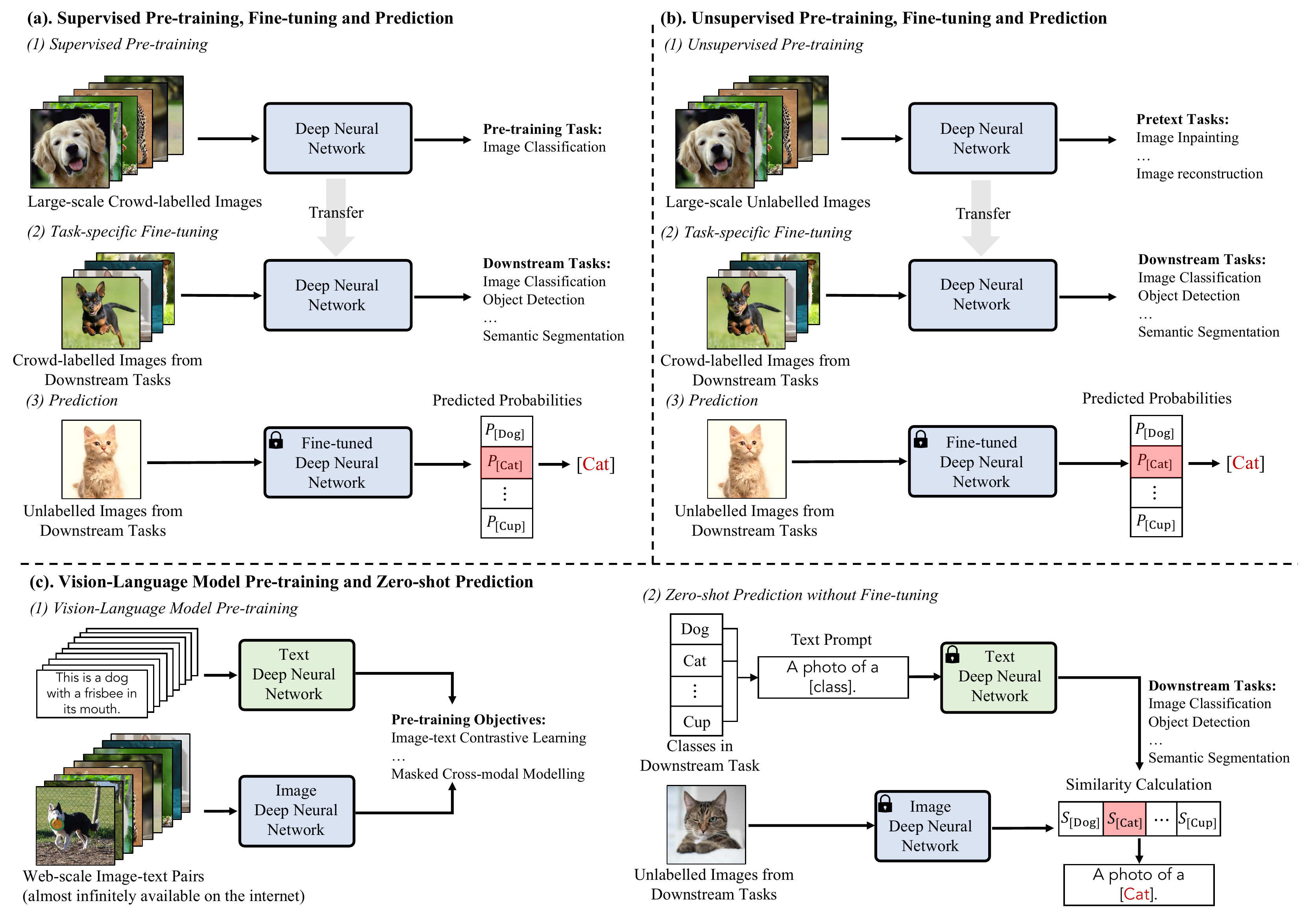}
    \caption{
    Three DNN training paradigms in visual recognition. Compared with the paradigms in \textbf{(a)} and \textbf{(b)} that requires fine-tuning for each specific task with task-specific labelled data, the new learning paradigm with VLMs in \textbf{(c)} enables effective usage of web data and zero-shot predictions without task-specific fine-tuning.
    }
    \label{fig.intro}
\end{figure*}

Following the great success of \textit{Vision-Language Model Pre-training and Zero-shot Prediction}, two lines of research have been intensively investigated beyond various VLM pre-training studies.
The first line explores VLMs with transfer learning~\cite{zhou2022learning,zhou2022conditional,gao2021clip,zhang2021tip}. It is evidenced by several transfer approaches, $e.g.$, prompt tuning~\cite{zhou2022learning,zhou2022conditional}, visual adaptation~\cite{gao2021clip,zhang2021tip}, etc., all sharing the same target for effective adaptation of pre-trained VLMs towards various downstream tasks. 
The second line explores VLMs with knowledge distillation~\cite{ding2022decoupling,gu2021open, du2022learning}, $e.g.$, several studies~\cite{ding2022decoupling,gu2021open, du2022learning} explore how to distill knowledge from VLMs to downstream tasks, aiming for better performance in object detection, semantic segmentation, etc.

Despite the intensive interest in harvesting the vast knowledge from VLMs as evidenced by a great number of recent papers as shown in Fig.~\ref{fig:trend}, the research community is short of a comprehensive survey that can help sort out existing VLM-based visual recognition studies, the facing challenges, as well as future research directions.
We aim to fill up this gap by performing a systematic survey of VLM studies in various visual recognition tasks including image classification, object detection, semantic segmentation, etc. We conduct the survey from different perspectives including background, foundations, datasets, technical approaches, benchmarking, and future research directions. We believe that this survey will provide a clear big picture on what we have achieved, and we could further achieve along this emerging yet very prospective research direction.

In summary, the main contributions of this work are threefold. 
    \textit{First}, it presents a systematic review of 
    VLMs for visual recognition tasks including image classification, object detection and semantic segmentation. To the best of our knowledge, this is the \textit{first} survey of VLMs 
    for visual recognition, which provides a big picture of this promising research filed with comprehensive summary and categorization of existing studies.
    \textit{Second}, it studies the up-to-date progress of VLMs for visual recognition, including a comprehensive benchmarking and discussion of existing work over multiple public datasets.
    \textit{Third}, it shares several research challenges and potential research directions that could be pursued in VLMs 
    for visual recognition.

The rest of this survey is organized as follows. 
Section \ref{Background} introduces the paradigm development of visual recognition and several related surveys. 
Section \ref{Foundation} describes the foundations of VLMs, including widely used deep network architectures, pre-training objectives, pre-training frameworks and downstream tasks in VLM evaluations. 
Section \ref{Dataset} introduces the commonly used datasets in 
VLM pre-training and evaluations. 
Section \ref{Sec.VLP} reviews and categorizes 
VLM pre-training methods. 
Sections \ref{sec:TL} and \ref{Sec.KD} provide a systematic review of transfer learning and knowledge distillation approaches for VLMs
, respectively. 
Section \ref{sec.benchmark} benchmarks 
the reviewed methods on multiple widely-adopted datasets. Finally, we share several promising VLM research directions 
in Section \ref{sec.future}.

\section{Background}\label{Background}

This section first presents the development of the training paradigm of visual recognition and how it evolves towards the paradigm \textit{Vision-Language Model Pre-training and Zero-shot Prediction}.
Then, we introduce the development of the vision-language models (VLMs) for visual recognition.
We also discuss several related surveys to highlight the scope and contributions of this survey.

\subsection{Training Paradigms for Visual Recognition}

The development of visual recognition paradigms can be broadly divided into five stages, including (1) \textit{Traditional Machine Learning and Prediction}, (2) \textit{Deep Learning from Scratch and Prediction}, (3) \textit{Supervised Pre-training, Fine-tuning and Prediction}, (4) \textit{Unsupervised Pre-training, Fine-tuning and Prediction} and
(5) \textit{Vision-language Model Pre-training and Zero-shot Prediction}. 
In what following, we introduce, compare and analyze the five training paradigms in detail.

\subsubsection{Traditional Machine Learning and Prediction}

Before the deep learning era~\cite{krizhevsky2017imagenet}, visual recognition studies rely heavily on \textit{feature engineering} with hand-crafted features~\cite{mathur2008multiclass,lowe2004distinctive} and lightweight learning models ~\cite{breiman2001random,peterson2009k,cortes1995support} that classify the hand-crafted features into pre-defined semantic categories.
However, this paradigm requires domain experts for crafting effective features for specific visual recognition tasks, which does not cope with complex tasks well and also has poor scalability.

\subsubsection{Deep Learning from Scratch and Prediction}

With the advent of deep learning~\cite{krizhevsky2017imagenet,simonyan2014very,he2016deep}, visual recognition research has achieved great success by leveraging end-to-end trainable DNNs that circumvent the complicated \textit{feature engineering} and allow focusing on the \textit{architecture engineering} of neural networks for learning effective features. For example, ResNet~\cite{he2016deep} enables very deep networks by a skip design and allows learning from massive crowd-labelled data with unprecedented performance on the challenging ImageNet benchmark~\cite{deng2009imagenet}. However, the turn from traditional machine learning toward deep learning raises two new grand challenges: the slow convergence of DNN training under the classical setup of \textit{Deep Learning from Scratch} and the laborious collection of large-scale, task-specific, and crowd-labelled data~\cite{radford2021learning} in DNN training.

\begin{figure}[ht]
    \centering
    \includegraphics[width=0.48\textwidth]{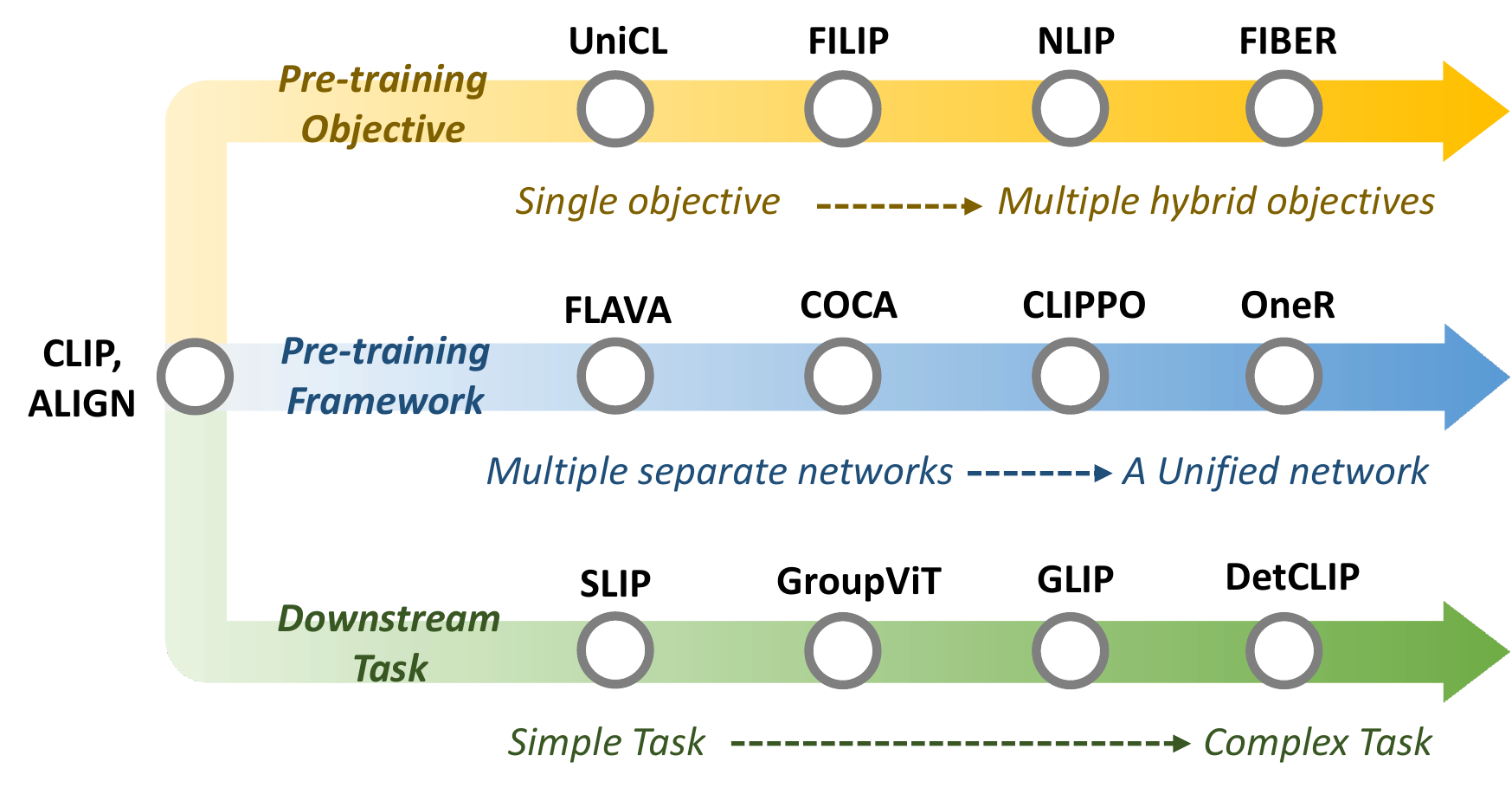}
    \caption{Illustration of development of VLMs for visual recognition.}
    \label{fig:development}
\end{figure}

\subsubsection{Supervised Pre-training, Fine-tuning and Prediction}

With the discovery that features learned from labelled large-scale datasets can be transferred to downstream tasks~\cite{girshick2015fast}, the paradigm \textit{Deep Learning from Scratch and Prediction} has been gradually replaced by a new paradigm of \textit{Supervised Pre-training, Fine-tuning and Prediction}.
This new learning paradigm, as illustrated in Fig.~\ref{fig.intro} {(a)}, pre-trains DNNs on large-scale labelled data ($e.g.$, ImageNet) with a supervised
loss and then fine-tunes the pre-trained DNN with task-specific training data~\cite{girshick2015fast}.
As the pre-trained DNNs have learned certain visual knowledge, it can accelerate network convergence and help train well-performing models with limited task-specific training data.

\subsubsection{Unsupervised Pre-training, Fine-tuning \& 
Prediction}

Though \textit{Supervised Pre-training, Fine-tuning and Prediction} achieves state-of-the-art performance on many visual recognition tasks, it requires large-scale labelled data in pre-training.
To mitigate this constraint, \cite{he2020momentum,chen2020simple} adopt a new learning paradigm \textit{Unsupervised Pre-training, Fine-tuning and Prediction} that explores self-supervised learning to learn useful and transferable representations from unlabelled data,
as illustrated in Fig.~\ref{fig.intro} {(b)}.
To this end, various self-supervised training objectives~\cite{he2022masked,he2020momentum} have been proposed 
including 
masked image modelling that models cross-patch relations~\cite{he2022masked}, contrastive learning that learns discriminative features by contrasting training samples~\cite{he2020momentum}, etc.
The self-supervised pre-trained models are then fine-tuned on downstream tasks with labelled task-specific training data.
Since this paradigm does not require labelled data in pre-training, it 
can exploit more training
data for learning useful and transferable features, leading to even better performance as compared with the supervised pre-training~\cite{chen2020simple,he2020momentum}.

\begin{figure*}[!t]
    \centering
    \includegraphics[width=0.88\textwidth]{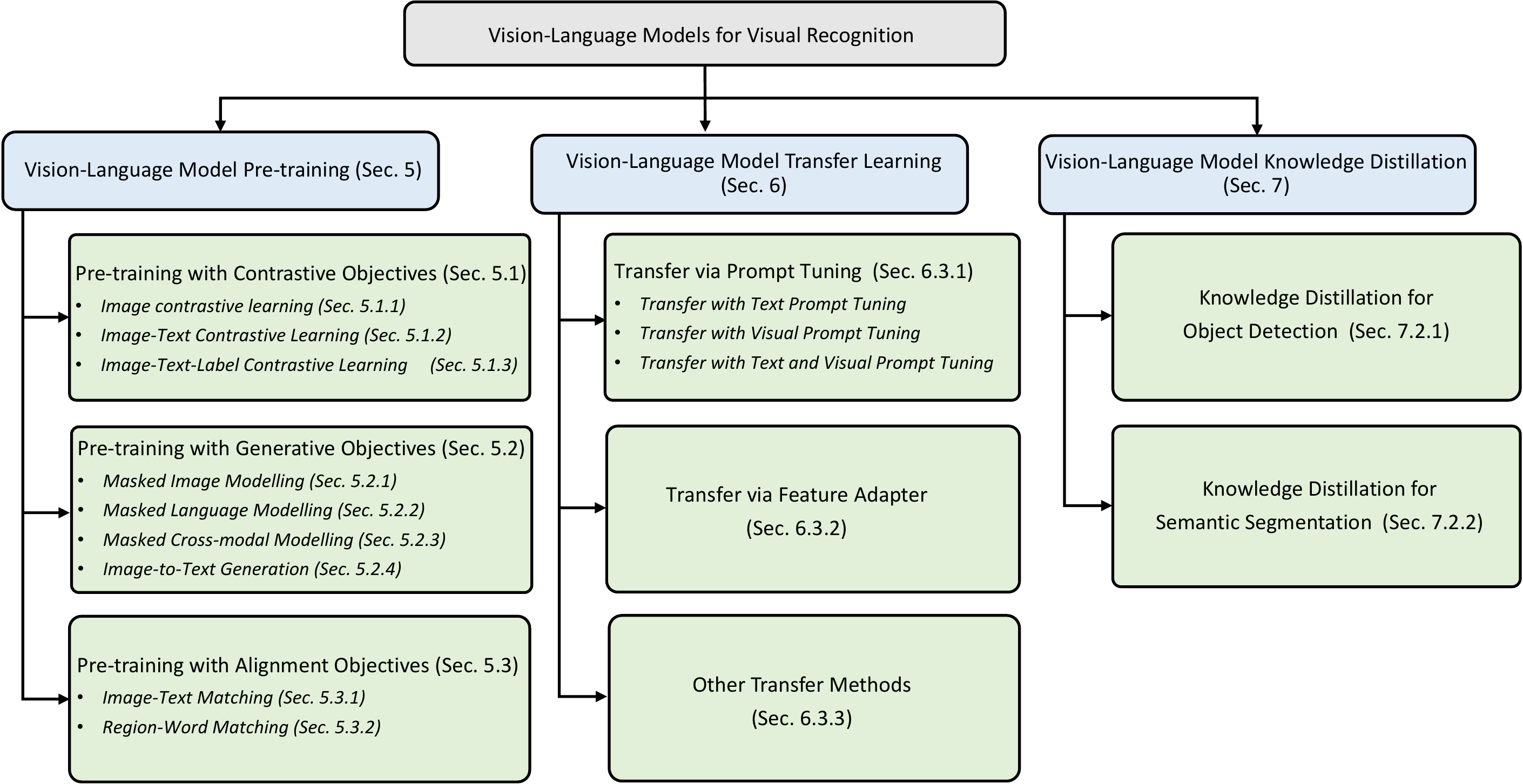}
    \caption{ Typology of vision-language models for visual recognition.}
    \label{fig:arch}
\end{figure*}

\subsubsection{VLM Pre-training and Zero-shot Prediction}

Though \textit{Pre-training and Fine-tuning}
with either supervised or unsupervised pre-training 
improves the network convergence, 
it still requires a fine-tuning stage with labelled task data as shown in Figs.~\ref{fig.intro} {(a)} and {(b)}.
Motivated by great success in natural language processing~\cite{devlin2018bert,radford2018improving,radford2019language}, a new deep learning paradigm named \textit{Vision-Language Model Pre-training and Zero-shot Prediction} has been proposed for visual recognition, as shown in Fig.~\ref{fig.intro} {(c)}.
With large-scale image-text pairs that are almost infinitely available on the internet, a VLM is pre-trained by certain vision-language objectives~\cite{radford2021learning,yao2021filip,yu2022coca} which captures rich vision-language knowledge and can perform zero-shot predictions (without fine-tuning) on downstream visual recognition tasks by matching the embeddings of any given images and texts.

Compared with \textit{Pre-training and Fine-tuning}, this new paradigm enables effective use of large-scale web data and zero-shot predictions without task-specific fine-tuning. Most existing research attempts to improve VLMs from 3 perspectives: 1) collecting large-scale informative image-text data, 2) designing high-capacity models for effective learning from big data, 3) designing new pre-training objectives for learning effective VLMs.  
In this paper, we provide a systematic survey of this new vision-language learning paradigm aiming to provide a clear big picture on exiting VLM studies, the facing challenges and future directions for this challenging but promising research filed.

\subsection{Development of VLMs for Visual Recognition}

Visual recognition related VLM studies have made great progresses since the development of CLIP~\cite{radford2021learning}.
We present VLMs for visual recognition from three aspects as illustrated in Fig.~\ref{fig:development}: (1) \textit{Pre-training objectives: from ``a single objective'' to ``multiple hybrid objectives''.} Early VLMs~\cite{radford2021learning,jia2021scaling} generally adopt a single pre-training objective, whereas recent VLMs~\cite{singh2022flava,yao2021filip} introduce multiple objectives ($e.g.$, contrastive, alignment and generative objectives) for exploring their synergy for more robust VLMs and better performance in downstream tasks; (2) \textit{Pre-training frameworks: from ``multiple separate networks'' to ``a unified network''.} Early VLMs~\cite{radford2021learning, jia2021scaling} employ two-tower pre-training frameworks, whereas recent VLMs~\cite{tschannen2022image,jang2023unifying} attempt one-tower pre-training framework that encodes images and texts with a unified network with less GPU memory usage yet more efficient communications across data modalities; 3) \textit{Downstream tasks: from simple to complex tasks.} Early VLMs~\cite{radford2021learning,jia2021scaling} focus on image-level visual recognition tasks, whereas recent VLMs~\cite{yaodetclip,luo2022segclip} are more general-purpose which can also work for dense prediction tasks that are complex and require localization related knowledge.

\subsection{Relevant Surveys}

To the best of our knowledge, this is the \textit{first} survey that reviews VLMs  
for various visual recognition tasks. 
Several relevant surveys have been conducted which 
review VLMs for vision-language tasks instead such as visual question answering~\cite{antol2015vqa}, natural language for visual reasoning~\cite{suhr2018corpus} 
, and phrase grounding~\cite{karpathy2014deep}. 
For instance, Li \textit{et al.}~\cite{li2022vision} shared advances on vision-language tasks, including VLM 
pre-training for various task-specific methods.
Du \textit{et al.}~\cite{du2022survey} and Chen \textit{et al.}~\cite{chen2023vlp} reviewed  
VLM pre-training for vision-language tasks~\cite{antol2015vqa,suhr2018corpus,karpathy2014deep}. 
Xu \textit{et al.}~\cite{xu2022multimodal} and Wang \textit{et al.}~\cite{wang2023large} shared recent progress of multi-modal learning on multi-modal tasks.
Differently, we review VLMs for visual recognition tasks from three major aspects: 1) Recent progress of VLM pre-training for visual recognition tasks; 2) Two typical transfer approaches from VLMs to visual recognition tasks;  
3) Benchmarking of VLM pre-training methods on visual recognition tasks.

\section{VLM Foundations}\label{Foundation}

VLM pre-training~\cite{radford2021learning,jia2021scaling} aims to pre-train a VLM to learn image-text correlation, targeting effective zero-shot predictions on visual recognition tasks~\cite{he2016deep,ren2015faster,chen2017deeplab}. Given image-text pairs~\cite{schuhmann2021laion,schuhmann2022laion}, it first employs a text encoder and an image encoder to extract image and text features~\cite{he2016deep,dosovitskiy2020image,vaswani2017attention,devlin2018bert} and then learns the vision-language correlation with certain pre-training objectives~\cite{radford2021learning,jia2021scaling}.
Hence, VLMs can be evaluated on unseen data in a zero-shot manner~\cite{radford2021learning,jia2021scaling} by matching the embeddings of any given images and texts. 
This section introduces the foundations of VLM pre-training, including common network architectures for extracting image and text features, pre-training objectives for modelling vision-language correlation, 
frameworks for VLM pre-training
and downstream tasks for VLM evaluations.

\subsection{Network Architectures}

VLM pre-training works with a deep neural network that extracts image and text features from $N$ image-text pairs within a pre-training dataset $\mathcal{D} = \{ x_n^I, x_n^T \}_{n=1}^N$, where $x_n^I$ and $x_n^T$ denote an image sample and its paired text sample. The deep neural network has an image encoder $f_{\theta}$ and a text encoder $f_{\phi}$, which encode the image and text (from an image-text pair \{$x_n^I$, $x_n^T$\}) into an image embedding $z_n^I = f_{\theta}(x_n^I)$ and a text embedding $z_n^T = f_{\phi}(x_n^T)$, respectively. This section presents the architecture of widely-adopted deep neural networks in VLM pre-training.

\subsubsection{Architectures for Learning Image Features}

Two types of network architectures have been widely adopted to learn image features, namely, CNN-based architectures and Transformer-based architectures.

\noindent \textbf{CNN-based Architectures.} 
Different ConvNets 
($e.g.$, VGG~\cite{simonyan2014very}, ResNet~\cite{he2016deep} and EfficientNet~\cite{tan2019efficientnet}) have been designed for learning image features. 
Being one of the most popular ConvNet in VLM pre-training, ResNet~\cite{he2016deep} adopts skip connections between convolution blocks which mitigates gradient vanishing and explosion and enables very deep neural networks.   
For better feature extraction and vision-language 
modelling, several studies~\cite{radford2021learning} modify the original network architecture~\cite{he2016deep,tan2019efficientnet}. Take ResNet as an example. They introduce the ResNet-D~\cite{he2019bag}, employ the antialiased rect-2 blur pooling in~\cite{zhang2019making}, and replace the global average pooling with an attention pooling in the transformer multi-head  attention~\cite{vaswani2017attention}. 

\noindent \textbf{Transformer-based Architectures.}
Transformers have recently been extensively explored in visual recognition tasks, such as image classification~\cite{dosovitskiy2020image}, object detection~\cite{carion2020end} and semantic segmentation~\cite{xie2021segformer}. 
As a standard Transformer architecture for image feature learning, ViT~\cite{dosovitskiy2020image} employs a stack of Transformer blocks each of which consists of a multi-head self-attention layer and a feed-forward network. 
The input image is first split into fixed-size patches 
and then fed to the Transformer encoder after linear projection and position embedding. 
\cite{radford2021learning,mu2022slip,yao2021filip} modify ViT by adding a normalization layer before the transformer encoder.

\subsubsection{Architectures for Learning Language Features}
Transformer \& its variants~\cite{vaswani2017attention,radford2019language,devlin2018bert} have been widely adopted for learning text features. 
The standard Transformer~\cite{vaswani2017attention} has an encoder-decoder structure, where the encoder has 6 blocks each of which has a multi-head self-attention layer and a multi-layer perceptron (MLP). The decoder also has 6 blocks each of which has 
a multi-head attention layer, a masked multi-head layer and a MLP. Most VLM studies such as CLIP~\cite{radford2021learning} adopt the standard Transformer~\cite{vaswani2017attention} with minor modifications as in GPT$_{2}$~\cite{radford2019language}, and train from scratch without initialization with GPT$_{2}$ weights.

\subsection{VLM Pre-training Objectives}
\label{sec:pre_obj}

As the core of VLM, various vision-language pre-training objectives~\cite{radford2021learning,he2020momentum,yang2022unified,devlin2018bert,he2021masked,yu2022coca,singh2022flava,li2022grounded} have been designed for learning rich vision-language correlation. They fall broadly into three categories: 
contrastive objectives, generative objectives and alignment objectives.

\subsubsection{Contrastive Objectives} 
\label{sec.3.2.1}

Contrastive objectives train VLMs to learn discriminative representations by pulling paired samples close and pushing others faraway in the feature space~\cite{radford2021learning,he2020momentum,yang2022unified}.

\noindent \textbf{Image Contrastive Learning} aims to learn discriminative image features~\cite{he2020momentum,chen2020simple} by forcing a query image to be close with its positive keys ($i.e.$, its data augmentations) and faraway from its negative keys ($i.e.$, other images) in the embedding space.  
Given a batch of $B$ images, contrastive-learning objectives ($e.g.$, InfoNCE~\cite{oord2018representation} and its variants~\cite{chen2020simple,he2020momentum}) are usually formulated as follows:
\begin{equation}
    \mathcal{L}_I^{\mathrm{InfoNCE}}=- \frac{1}{B} \sum_{i=1}^B \log \frac{\exp{(z_i^I\cdot z^I_+/\tau)}}{\sum_{j=1, j \neq i}^{B + 1}{\exp(z_i^I\cdot z^I_j/\tau)}},
\label{eq:infonce}
\end{equation}
where $z_{i}^I$ is the query embedding, $\{z_j^I\}_{j=1,j \neq i}^{B + 1}$ are key embeddings, where $z^I_+$ stands for $z_{i}^I$'s positive key and the rest are $z_{i}^I$'s negative keys.
$\tau$ is a temperature hyper-parameter that controls the density of the learned representation.

\noindent \textbf{Image-Text Contrastive Learning} aims to learn discriminative image-text representations by pulling the embeddings of paired images and texts close while pushing others~\cite{radford2021learning,jia2021scaling} away. 
It is usually achieved by minimizing a symmetrical image-text infoNCE loss~\cite{radford2021learning}, $i.e.$, $\mathcal{L}_{\mathrm{infoNCE}}^{IT} = \mathcal{L}_{I \rightarrow T} + \mathcal{L}_{T \rightarrow I}$, where $ \mathcal{L}_{I \rightarrow T}$ contrasts the query image with the text keys 
while $ \mathcal{L}_{T \rightarrow I}$ contrasts the query text with image keys. Given a batch of $B$ image-text pairs, $ \mathcal{L}_{I \rightarrow T}$ and  $ \mathcal{L}_{T \rightarrow I}$ are defined as follows:

\begin{equation}
    \mathcal{L}_{I \rightarrow T}=- \frac{1}{B} \sum_{i=1}^B \log \frac{\exp{(z_i^I\cdot z^T_i/\tau)}}{\sum_{j=1}^{B}{\exp(z^I_i\cdot z^T_j/\tau)}},
    \label{eq_img2text}
\end{equation}
\begin{equation}
    \mathcal{L}_{T \rightarrow I}=- \frac{1}{B} \sum_{i=1}^B \log \frac{\exp{(z_i^T\cdot z^I_i/\tau)}}{\sum_{j=1}^{B}{\exp(z^T_i\cdot z^I_j/\tau)}},
    \label{eq_text2img}
\end{equation}
where $z^I$ and $z^T$ stand for the image embeddings and text embeddings, respectively.

\noindent \textbf{Image-Text-Label Contrastive Learning.} Image-text-label contrastive learning~\cite{yang2022unified} introduces Supervised Contrastive Learning~\cite{khosla2020supervised} into image-text contrastive
learning, which is defined by reformulating Eqs.~\ref{eq_img2text} and~\ref{eq_text2img} as follows:
\begin{equation}
    \mathcal{L}_{I \rightarrow T}^{ITL}=- \sum_{i=1}^B \frac{1}{|\mathcal{P}(i)|} \sum_{k\in \mathcal{P}(i)} \log \frac{\exp{(z_i^I\cdot z^T_k/\tau)}}{\sum_{j=1}^{B}{\exp(z^I_i\cdot z^T_j/\tau)}},
    \label{eq_img2text2label}
\end{equation}
\begin{equation}
    \mathcal{L}_{T \rightarrow I}^{ITL}=- \sum_{i=1}^B \frac{1}{|\mathcal{P}(i)|} \sum_{k\in \mathcal{P}(i)} \log \frac{\exp{(z^T_i \cdot z_k^I /\tau)}}{\sum_{j=1}^{B}{\exp(z^T_i \cdot z^I_j /\tau)}},
    \label{eq_text2img2label}
\end{equation}
where $k \in \mathcal{P}(i) = \{k|k \in B, y_k = y_i \}$~\cite{yang2022unified} and $y$ is the category label of $(z^{I}, z^{T})$.
With Eqs.~\ref{eq_img2text2label} and~\ref{eq_text2img2label}, the image-text-label infoNCE loss is defined as: $\mathcal{L}^{ITL}_{\mathrm{infoNCE}} = \mathcal{L}_{I \rightarrow T}^{ITL} + \mathcal{L}_{T \rightarrow I}^{ITL}$.

\subsubsection{Generative Objectives}

Generative objectives learn 
semantic features by training networks to generate image/text data via image generation~\cite{he2020momentum,bao2021beit}, language generation~\cite{devlin2018bert,yu2022coca}, or cross-modal generation~\cite{singh2022flava}.

\noindent \textbf{Masked Image Modelling} learns cross-patch correlation by masking and reconstructing images~\cite{he2022masked,bao2021beit}. It masks a set of patches of an input image randomly and trains the encoder to reconstruct the masked patches conditioned on unmasked patches. 
Given a batch of $B$ images, the loss function can be formulated as:
\begin{equation}
    \mathcal{L}_{MIM} = - \frac{1}{B} \sum_{i=1}^B \log f_{\theta} ( \ \overline{x}^I_i \ | \ \hat{x}^I_i \ ),
    \label{eq:MIM}
\end{equation}
where $\overline{x}^I_i$ and $\hat{x}^I_i$ denote the masked patches and the unmasked patches in $x^I_i$, respectively.

\noindent \textbf{Masked Language Modelling} is a widely adopted pre-training objective in NLP~\cite{devlin2018bert}. It randomly masks a certain percentage ($e.g.$, 15\% in BERT~\cite{devlin2018bert}) of the input text tokens, and reconstruct them with unmasked tokens: 
\begin{equation}
    \mathcal{L}_{MLM} = - \frac{1}{B} \sum_{i=1}^B \log f_{\phi} ( \ \overline{x}^T_i \ | \ \hat{x}^T_i \ ),
    \label{eq:MLM}
\end{equation}
where $\overline{x}^T_i$ and $\hat{x}^T_i$ denote the masked and unmasked tokens in $x^T_i$, respectively. $B$ denotes the batch size.

\noindent \textbf{Masked Cross-Modal Modelling}
integrates masked image modelling and masked language modelling~\cite{singh2022flava}.
Given an image-text pair, it randomly masks a subset of image patches and a subset of text tokens and then learns to reconstruct them conditioned on unmasked image patches and unmasked text tokens as follows: 
\begin{equation}
    \mathcal{L}_{MCM} = - \frac{1}{B} \sum_{i=1}^B [ \log f_{\theta} (  \overline{x}^I_i  |  \hat{x}^I_i, \hat{x}^T_i ) + \log f_{\phi} (  \overline{x}^T_i  |  \hat{x}^I_i, \hat{x}^T_i )],
\label{eq:MMM}
\end{equation}
where $\overline{x}^I_i$/$\hat{x}^I_i$ denotes the masked/unmasked patches in $x^I_i$, $\overline{x}^T_i$/$\hat{x}^T_i$ denotes the masked/unmasked text tokens in $x^T_i$.

\noindent \textbf{Image-to-Text Generation} aims to predict text $x^T$ autoregressively based on the image paired with $x^T$~\cite{yu2022coca}:

\begin{equation}
    \mathcal{L}_{ITG} = - \sum^L_{l=1} \ \log \ f_{\theta} (x^T \ | \ x^T_{<l}, z^I),
    \label{eq:caption}
\end{equation}
where $L$ denotes the number of tokens to be predicted for $x^T$ and $z^{I}$ is the embedding of the image paired with $x^T$.

\subsubsection{Alignment Objectives}

Alignment objectives align the image-text pair via global image-text matching~\cite{doucoarse,bao2021vlmo} or local region-word matching~\cite{li2022grounded,yaodetclip} on embedding space.

\noindent \textbf{Image-Text Matching}
models global correlation between images and texts~\cite{doucoarse,bao2021vlmo}, which can be formulated with a score function $\mathcal{S}(\cdot)$ that measures the alignment probability between the image and text and a binary classification loss:
\begin{equation}
\mathcal{L}_{IT} =   p \log  \mathcal{S} (z^I, z^T) + (1-p)  \log  (1 - \mathcal{S} (z^I, z^T)),
\label{eq:ITM}
\end{equation}
where $p$ is 1 if the image and text are paired and 0 otherwise. 

\noindent \textbf{Region-Word Matching}
aims to model local cross-modal correlation ($i.e.$, between ``image regions" and ``words") in image-text pairs~\cite{li2022grounded,yaodetclip} for dense visual recognition tasks such as object detection. It can be formulated as:
\begin{equation}
\mathcal{L}_{RW} =  p  \log  \mathcal{S}^r (r^I, w^T) + (1-p)  \log  (1 - \mathcal{S}^r (r^I, w^T)),
\label{eq:RWM}
\end{equation}
where $(r^I, w^T)$ denotes a region-word pair and $p = 1$ if the region and word are paired otherwise $p = 0$. $\mathcal{S}^r (\cdot)$ denotes a local score function that measures the similarity between ``image regions" and ``words".

\subsection{VLM Pre-training Frameworks}

This section presents widely adopted frameworks in VLM pre-training, including two-tower, two-leg and one-tower pre-training frameworks.

\begin{figure}[ht]
    \centering
    \includegraphics[width=0.5\textwidth]{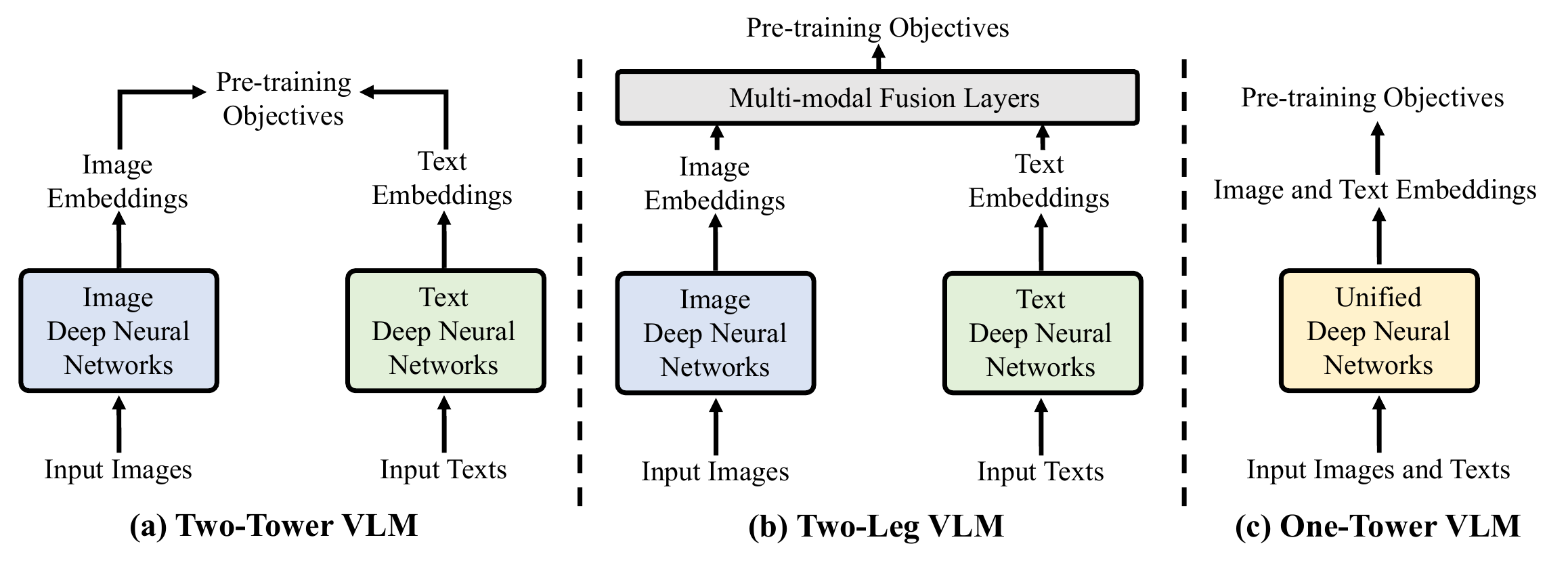}
    \caption{Illustration of typical VLM pre-training frameworks.}
    \label{fig:tower}
\end{figure}

Specifically, two-tower framework has been widely adopted in VLM pre-training~\cite{radford2021learning,jia2021scaling}, where input images and texts are encoded with two separate encoders respectively, as illustrated in Fig.~\ref{fig:tower} (a).
Slightly differently, two-leg framework~\cite{singh2022flava,yu2022coca} introduces additional multi-modal fusion layers which enable feature interaction between image and text modalities, as illustrated in Fig.~\ref{fig:tower} (b).
As a comparison, one-tower VLMs~\cite{tschannen2022image,jang2023unifying} attempt to unify vision and language learning in a single encoder as illustrated in Fig.~\ref{fig:tower} (c), aiming to facilitate efficient communications across data modalities.

\subsection{Evaluation Setups and Downstream Tasks}\label{sec.eval}
This section presents widely adopted setups and downstream tasks in VLM evaluation. The setups include \textit{zero-shot prediction} and \textit{linear probing}, and the downstream tasks include image classification, object detection, semantic segmentation, image-text retrieval, and action recognition.

\renewcommand\arraystretch{0.95}
\begin{table*}[t]
    \centering
    \caption{Summary of the widely used image-text datasets for VLM pre-training. \textcolor{magenta}{[link]} directs to dataset websites.}
    \begin{tabular}{l|c|c|c|c}
        \toprule[1.pt]
        \textbf{Dataset} & \textbf{Year} & \textbf{Num. of Image-Text Pairs} & \textbf{Language} & \textbf{Public}\\
        \midrule
        SBU Caption~\cite{ordonez2011im2text}~\href{https://www.cs.rice.edu/~vo9/sbucaptions/}{[link]} & 2011 &  1M & English & \cmark\\
        COCO Caption~\cite{chen2015microsoft}~\href{https://github.com/tylin/coco-caption}{[link]} & 2016 & 1.5M & English &\cmark\\
        Yahoo Flickr Creative Commons 100 Million (YFCC100M)~\cite{thomee2016yfcc100m}~\href{http://projects.dfki.uni-kl.de/yfcc100m/}{[link]} & 2016 & 100M & English&\cmark\\
        Visual Genome (VG)~\cite{krishna2017visual}~\href{http://visualgenome.org/}{[link]} & 2017 & 5.4 M & English&\cmark\\
        Conceptual Captions (CC3M)~\cite{sharma2018conceptual}~\href{https://ai.google.com/research/ConceptualCaptions/}{[link]} & 2018 & 3.3M & English&\cmark\\
        Localized Narratives (LN)~\cite{pont2020connecting}~\href{https://google.github.io/localized-narratives/}{[link]} & 2020 & 0.87M & English&\cmark\\
        Conceptual 12M (CC12M)~\cite{changpinyo2021conceptual}~\href{https://github.com/google-research-datasets/conceptual-12m}{[link]} & 2021 & 12M & English&\cmark\\
        Wikipedia-based Image Text (WIT)~\cite{srinivasan2021wit}~\href{https://github.com/google-research-datasets/wit}{[link]} & 2021 & 37.6M & 108 Languages&\cmark \\
        Red Caps (RC)~\cite{desai2021redcaps}~\href{https://redcaps.xyz/}{[link]} & 2021 & 12M & English&\cmark\\
        LAION400M~\cite{schuhmann2021laion}~\href{https://laion.ai/blog/laion-400-open-dataset/}{[link]} & 2021 & 400M & English&\cmark\\
        LAION5B~\cite{schuhmann2022laion}~\href{https://laion.ai/blog/laion-5b/}{[link]} & 2022 & 5B & Over 100 Languages&\cmark\\
        
        WuKong~\cite{gu2022wukong}~\href{https://wukong-dataset.github.io/wukong-dataset/}{[link]} & 2022 & 100M & Chinese &\cmark\\
        \midrule
        CLIP~\cite{radford2021learning} & 2021 & 400M & English & \xmark \\
        ALIGN~\cite{jia2021scaling} &2021 &1.8B & English & \xmark \\
        FILIP~\cite{yao2021filip} & 2021&300M & English & \xmark \\
        WebLI~\cite{chen2022pali} & 2022 & 12B & 109 Languages&\xmark\\
        \bottomrule[1.pt]
    \end{tabular}
    \label{tab.pretrain_datasets}
\end{table*}

\subsubsection{Zero-shot Prediction}

As the most common way of evaluating VLMs' generalization capability~\cite{radford2021learning,jia2021scaling,yao2021filip,leeuniclip,mu2022slip}, zero-shot prediction directly applies pre-trained VLMs to downstream tasks without any task-specific fine-tuning~\cite{radford2021learning}.

\noindent\textbf{Image Classification}~\cite{he2016deep,simonyan2014very} aims to classify images into pre-defined categories. VLMs achieve zero-shot image classification by comparing the embeddings of images and texts, where ``prompt engineering" is often employed to generate task-related prompts like ``a photo of a [\verb|label|]."~\cite{radford2021learning}.

\noindent \textbf{Semantic Segmentation}~\cite{chen2017deeplab} aims to assign a category label to each pixel in images. Pre-trained VLMs achieve zero-shot prediction for segmentation tasks by comparing the embeddings of the given image pixels and texts.

\noindent \textbf{Object Detection}~\cite{girshick2015fast,ren2015faster} aims to localize and classify objects in images, which is important for various vision applications. With the object locating ability learned from auxiliary datasets~\cite{shao2019objects365,kamath2021mdetr}, pre-trained VLMs achieve zero-shot prediction for object detection tasks by comparing the embeddings of the given object proposals and texts.

\noindent \textbf{Image-Text Retrieval}~\cite{cao2022image} aims to retrieve the demanded samples from one modality given the cues from another modality, which consists of two tasks, $i.e.$, text-to-image retrieval that retrieves images based on texts and image-to-text retrieval that retrieves texts based on images.

\subsubsection{Linear Probing}

Linear probing has been widely adopted in VLM evaluations~\cite{radford2021learning}. 
It freezes the pre-trained VLM and trains a linear classifier to classify the VLM-encoded embeddings to assess the VLM representations. Image classification~\cite{he2016deep,simonyan2014very} and action recognition~\cite{soomro2012ucf101,miech2020rareact} have been widely adopted in such evaluations, where video clips are often sub-sampled for efficient recognition in action recognition tasks~\cite{radford2021learning}.

\section{Datasets}\label{Dataset}

This section summarizes the commonly used datasets for VLM pre-training and evaluations, as detailed in Tables~\ref{tab.pretrain_datasets}-\ref{tab.eval_datasets}.

\renewcommand\arraystretch{0.95}
\begin{table*}[t]
    \centering
    \caption{Summary of the widely-used visual recognition datasets for VLM evaluation. \textcolor{magenta}{[link]} directs to dataset websites.}
    \begin{tabular}{l|l|c|c|c|c|c}
        \toprule[1pt]
        \textbf{Task}&\textbf{Dataset} & \textbf{Year} &\textbf{Classes} & \textbf{Training} & \textbf{Testing} & \textbf{Evaluation Metric} \\
        \midrule
        Image Classification &MNIST~\cite{lecun1998gradient}~\href{http://yann.lecun.com/exdb/mnist/}{[link]} &1998& 10 & 60,000 & 10,000 & Accuracy \\
        &Caltech-101~\cite{fei2004learning}~\href{https://data.caltech.edu/records/mzrjq-6wc02}{[link]} & 2004& 102 & 3,060 & 6,085 & Mean Per Class\\
        &PASCAL VOC 2007 Classification~\cite{everingham2010pascal}~\href{http://host.robots.ox.ac.uk/pascal/VOC/voc2007/}{[link]} & 2007 & 20 & 5,011 & 4,952 & 11-point mAP \\
        &Oxford 102 Folwers~\cite{nilsback2008automated}~\href{https://www.robots.ox.ac.uk/~vgg/data/flowers/102/}{[link]} &2008& 102 & 2,040 & 6,149 & Mean Per Class\\
        &CIFAR-10~\cite{krizhevsky2009learning}~\href{https://www.cs.toronto.edu/~kriz/cifar.html}{[link]} & 2009 & 10 & 50,000 & 10,000 & Accuracy \\
        &CIFAR-100~\cite{krizhevsky2009learning}~\href{https://www.cs.toronto.edu/~kriz/cifar.html}{[link]} & 2009 & 100 & 50,000 & 10,000 & Accuracy\\
        &ImageNet-1k~\cite{deng2009imagenet}~\href{https://www.image-net.org/}{[link]} &2009& 1000 & 1,281,167 & 50,000 & Accuracy\\
        &SUN397~\cite{xiao2010sun}~\href{https://vision.princeton.edu/projects/2010/SUN/}{[link]} & 2010& 397 & 19,850 & 19,850 & Accuracy\\
        &SVHN~\cite{netzer2011reading}~\href{http://ufldl.stanford.edu/housenumbers/}{[link]} &2011&10 &73,257&26,032&Accuracy\\
        &STL-10~\cite{coates2011analysis}~\href{https://cs.stanford.edu/~acoates/stl10/}{[link]} &2011& 10 & 1,000 & 8,000 & Accuracy\\
        &GTSRB~\cite{stallkamp2011german}~\href{https://www.kaggle.com/datasets/meowmeowmeowmeowmeow/gtsrb-german-traffic-sign}{[link]} &2011& 43 & 26,640 & 12,630 & Accuracy\\
        &KITTI Distance~\cite{geiger2012we} ~\href{https://github.com/harshilpatel312/KITTI-distance-estimation}{[link]}& 2012& 4 & 6,770 & 711 & Accuracy\\
        &IIIT5k~\cite{mishra2012scene}~\href{https://cvit.iiit.ac.in/research/projects/cvit-projects/the-iiit-5k-word-dataset}{[link]} &2012&36 &2,000&3,000&Accuracy\\
        &Oxford-IIIT PETS~\cite{parkhi2012cats}~\href{https://www.robots.ox.ac.uk/~vgg/data/pets/}{[link]} &2012 & 37 & 3,680 & 3,669 & Mean Per Class\\
        &Stanford Cars~\cite{krause2013collecting}~\href{http://ai.stanford.edu/~jkrause/cars/car_dataset.html}{[link]} & 2013 & 196 & 8,144 & 8,041 & Accuracy\\
        &FGVC Aircraft~\cite{maji2013fine}~\href{https://www.robots.ox.ac.uk/~vgg/data/fgvc-aircraft/}{[link]} &2013 & 100 & 6,667 & 3,333 & Mean Per Class\\
        &Facial Emotion Recognition 2013~\cite{goodfellow2013challenges}~\href{https://www.kaggle.com/competitions/challenges-in-representation-learning-facial-expression-recognition-challenge/data}{[link]} &2013& 8 & 32,140 & 3,574 & Accuracy\\
        &Rendered SST2~\cite{socher2013recursive}~\href{https://github.com/openai/CLIP/blob/main/data/rendered-sst2.md}{[link]} &2013& 2 & 7,792 & 1,821 & Accuracy\\

        &Describable Textures (DTD)~\cite{cimpoi2014describing}~\href{https://www.robots.ox.ac.uk/~vgg/data/dtd/}{[link]} & 2014 & 47 & 3,760 & 1,880 & Accuracy\\
        &Food-101~\cite{bossard2014food}~\href{https://www.kaggle.com/datasets/dansbecker/food-101}{[link]} & 2014 & 102 & 75,750 & 25,250 & Accuracy  \\
        &Birdsnap~\cite{berg2014birdsnap}~\href{https://thomasberg.org/}{[link]} & 2014 & 500 & 42,283 & 2,149 & Accuracy\\
        &RESISC45~\cite{cheng2017remote}~\href{https://pan.baidu.com/s/1mifR6tU?_at_=1679281159364#list/path=%2F}{[link]} &2017& 45 & 3,150 & 25,200 & Accuracy\\
        &CLEVR Counts~\cite{johnson2017clevr}~\href{https://cs.stanford.edu/people/jcjohns/clevr/}{[link]} &2017& 8 & 2,000 & 500 & Accuracy\\
        &PatchCamelyon~\cite{veeling2018rotation}~\href{https://github.com/basveeling/pcam}{[link]} &2018& 2 & 294,912 & 32,768 & Accuracy\\

        &EuroSAT~\cite{helber2019eurosat}~\href{https://github.com/phelber/eurosat}{[link]} &2019& 10 & 10,000 & 5,000 & Accuracy\\
        
        &Hateful Memes~\cite{kiela2020hateful}~\href{https://ai.facebook.com/blog/hateful-memes-challenge-and-data-set/}{[link]} & 2020&2 & 8,500 & 500 & ROC AUC\\
        &Country211~\cite{radford2021learning}~\href{https://github.com/openai/CLIP/blob/main/data/country211.md}{[link]} &2021& 211 & 43,200 & 21,100 & Accuracy\\

        \midrule
        Image-Text Retrieval&Flickr30k~\cite{young2014image}~\href{https://shannon.cs.illinois.edu/DenotationGraph/}{[link]} &2014& - &31,783&-&Recall\\
        &COCO Caption~\cite{chen2015microsoft}~\href{https://github.com/tylin/coco-caption}{[link]}&2015 & - &82,783&5,000 &Recall\\
        \midrule
        
        Action Recognition&UCF101~\cite{soomro2012ucf101}~\href{https://www.crcv.ucf.edu/data/UCF101.php}{[link]} &2012& 101 & 9,537 & 1,794 & Accuracy\\
        &Kinetics700~\cite{carreira2019short}~\href{https://www.deepmind.com/open-source/kinetics}{[link]} &2019& 700 & 494,801 & 31,669 & Mean(top1, top5)\\
        &RareAct~\cite{miech2020rareact}~\href{https://github.com/antoine77340/RareAct}{[link]} &2020&122 &7,607&-&mWAP, mSAP\\

        \midrule
        
        Object Detection&COCO 2014 Detection~\cite{lin2014microsoft}~\href{https://www.kaggle.com/datasets/jeffaudi/coco-2014-dataset-for-yolov3}{[link]} & 2014& 80 & 83,000 & 41,000 & box mAP \\
        &COCO 2017 Detection~\cite{lin2014microsoft}~\href{https://www.kaggle.com/datasets/awsaf49/coco-2017-dataset}{[link]} & 2017 & 80 & 118,000 & 5,000 & box mAP \\
        &LVIS~\cite{gupta2019lvis}~\href{https://www.lvisdataset.org/}{[link]}& 2019& 1203 & 118,000 &5,000& box mAP \\
        &ODinW~\cite{li2022elevater}~\href{https://eval.ai/web/challenges/challenge-page/1839/overview}{[link]} &2022& 314 & 132413 & 20070 & box mAP \\ 
        \midrule
        Semantic Segmentation&PASCAL VOC 2012 Segmentation~\cite{everingham2010pascal}~\href{http://host.robots.ox.ac.uk/pascal/VOC/voc2012/}{[link]} & 2012&20&1464&1449&mIoU\\
        &PASCAL Content~\cite{mottaghi2014role}~\href{https://www.cs.stanford.edu/~roozbeh/pascal-context/}{[link]} & 2014 &459&4998&5105 &mIoU\\
        &Cityscapes~\cite{cordts2016cityscapes}~\href{https://www.cityscapes-dataset.com/}{[link]}& 2016 &19&2975&500&mIoU \\
        &ADE20k~\cite{zhou2017scene}~\href{https://groups.csail.mit.edu/vision/datasets/ADE20K/}{[link]}&2017&150& 25574 & 2000 & mIoU \\
        \bottomrule[1pt]
    \end{tabular}
    \label{tab.eval_datasets}
\end{table*}

\subsection{Datasets for Pre-training VLMs}

For VLM pre-training, multiple large-scale image-text datasets~\cite{radford2021learning,jia2021scaling,schuhmann2021laion,schuhmann2022laion} were collected from the internet. 
Compared with traditional crowd-labelled datasets~\cite{deng2009imagenet,cordts2016cityscapes,everingham2010pascal}, the image-text datasets~\cite{schuhmann2021laion,radford2021learning} are much larger and cheaper to collect.
For example, recent image-text datasets are generally at billion scale~\cite{schuhmann2021laion,schuhmann2022laion,chen2022pali}.
Beyond image-text datasets, several studies~\cite{li2022grounded,yu2022coca,yaodetclip,tschannen2022image} utilize auxiliary datasets to provide additional information for better vision-language modelling, $e.g.$, GLIP~\cite{li2022grounded} leverages Object365~\cite{shao2019objects365} for extracting region-level features.
The details of image-text datasets and auxiliary datasets for VLM pre-training are provided in Appendix B.

\subsection{Datasets for VLM Evaluation}

Many datasets have been adopted in VLM evaluations as shown in Table~\ref{tab.eval_datasets}, including $27$ for image classification, $4$ for object detection, $4$ for semantic segmentation, $2$ for image-text retrieval, and $3$ for action recognition (dataset details provided in Appendix C). For example, the $27$ image classification datasets cover a wide range of visual recognition tasks from fine-grained tasks like Oxford-IIIT PETS~\cite{parkhi2012cats} for pet identification and Stanford Cars~\cite{krause2013collecting} for car recognition, to general tasks like ImageNet~\cite{deng2009imagenet}.

\begin{table*}[!t]
    \setlength\tabcolsep{6pt}
    \centering
    \caption{Summary of vision-language model pre-training methods.
    Con: Contrastive Objective; Gen: Generative Objective; Align: Alignment Objective. $\dagger$, 
    $\ddagger$ and $\mathsection$ denote two-tower, two-leg and one-tower pre-training frameworks, respectively. $*$ denotes non-public datasets. \textcolor{magenta}{[code]} directs to code websites. }
    \resizebox{0.99\linewidth}{!}{
    \begin{tabular}{l|p{4.5cm}|p{1.7cm}|p{9.cm}}
    \toprule[1pt]
        \textbf{Method} & \textbf{Dataset}  & \textbf{Objective} & \textbf{Contribution} \\
        \midrule
        CLIP$\dagger$~\cite{radford2021learning}~\href{https://github.com/openai/CLIP}{[code]} & CLIP$*$ & Con & Propose image-text contrastive learning for VLM pre-training. \\
        ALIGN$\dagger$~\cite{jia2021scaling}  & ALIGN$*$  & Con & Leverage large-scale noisy data to scale-up VLM pre-training data.\\
        OTTER$\dagger$~\cite{wu2021data}~\href{https://github.com/facebookresearch/OTTER}{[code]} & CC3M, YFCC15M, WIT  & Con & Employ optimal transport for data efficient VLM pre-training.\\
        DeCLIP$\dagger$~\cite{li2021supervision}~\href{https://github.com/Sense-GVT/DeCLIP}{[code]} & CC3M, CC12M, YFCC100M, WIT$*$  & Con,Gen & Employ image/text self-supervision for data efficient VLM pre-training.  \\
        ZeroVL$\dagger$~\cite{cui2022contrastive}~\href{https://github.com/zerovl/ZeroVL}{[code]} & SBU, VG, CC3M, CC12M  & Con & Introduce data augmentation for data-efficient VLM pre-training.\\
        FILIP$\dagger$~\cite{yao2021filip} & FILIP$*$, CC3M, CC12M, YFCC100M & Con,Align & Leverage region-word similarity for fine-grained VLM pre-training.\\
        UniCL$\dagger$~\cite{yang2022unified}~\href{https://github.com/microsoft/UniCL}{[code]} & CC3M, CC12M, YFCC100M  & Con & Propose image-text-label contrastive learning for VLM pre-training.\\
        Florence$\dagger$~\cite{yuan2021florence} & FLD-900M$*$  & Con & 
        Scale up pre-training data and include depth and temporal information. 
        \\
        SLIP$\dagger$~\cite{mu2022slip} ~\href{https://github.com/facebookresearch/SLIP}{[code]} & YFCC100M  & Con & Introduce image self-supervision learning into VLM pre-training.\\
        PyramidCLIP$\dagger$~\cite{gao2022pyramidclip} & SBU, CC3M, CC12M, YFCC100M, LAION400M  & Con & Perform peer-level/cross-level contrastive learning within/across multiple semantic levels.\\
        ChineseCLIP$\dagger$~\cite{yang2022chinese}~\href{https://github.com/OFA-Sys/Chinese-CLIP}{[code]} & LAION5B, WuKong, VG, COCO  & Con & Collect large-scale Chinese image-text data and Introduce Chinese VLM.\\
        LiT$\dagger$~\cite{zhai2022lit}~\href{https://google-research.github.io/vision_transformer/lit/}{[project]} & CC12M, YFCC100M, WIT$*$  & Con & Propose contrastive tuning with the locked image encoder.\\
        AltCLIP$\dagger$~\cite{chen2022altclip}~\href{https://github.com/FlagAI-Open/FlagAI/tree/master/examples/AltCLIP}{[code]} & WuDao, LAION2B, LAION5B  & Con & Leverage the multilingual text encoder to achieve multilingual VLM. \\
        FLAVA$\ddagger$~\cite{singh2022flava}~\href{https://github.com/facebookresearch/multimodal/tree/main/examples/flava}{[code]} & COCO, SBU, LN, CC3M, VG, WIT, CC12M, RC, YFCC100M & Gen,Con,Align & Propose a universal and foundational VLM that tackles the single-modal ($i.e.$, image or text) and the multi-model cases at the same time. \\
        KELIP$\dagger$~\cite{ko2022large}~\href{https://github.com/navervision/KELIP}{[code]} & CUB200, WIT, YFCC15M, CC3M, CC12M, LAION400M, K-WIT$*$  & Con,Gen & Collect large-scale Korean image-text pair data and develop bilingual VLMs with Korean and English.\\
        COCA$\ddagger$~\cite{yu2022coca}~\href{https://github.com/lucidrains/CoCa-pytorch}{[code]} & ALIGN$*$  & Con,Gen & Combine contrastive learning and image captioning for pre-training.\\ 
        nCLIP$\dagger$~\cite{zhou2022non} & COCO, VG, SBU, CC3M, CC12M, YFCC14M  & Con,Align & Propose a non-contrastive pre-training objective ($i.e.$, a cross-entropy loss for global image-text matching) for VLM pre-training.\\ 
        K-lite$\dagger$~\cite{shen2022k}~\href{https://github.com/microsoft/klite}{[code]} & CC3M, CC12M, YFCC100M  & Con & Leverage auxiliary datasets for training transferable VLMs.\\
        NLIP$\ddagger$~\cite{huang2022nlip} & YFCC100M, COCO  & Con,Gen & Train noise-robust VLM via noise harmonization and completion.\\
        UniCLIP$\dagger$~\cite{leeuniclip} & CC3M, CC12M, YMCC100M  &Con & 
        Propose unified image-text and image-image contrastive learning.
        \\
        PaLI$\ddagger$~\cite{chen2022pali}~\href{https://ai.googleblog.com/2022/09/pali-scaling-language-image-learning-in.html}{[project]}  & WebLI$*$  & Gen & Scale up the data, model and language in VLM pre-taring. \\ 
        HiCLIP$\dagger$~\cite{geng2023hiclip} ~\href{https://github.com/jeykigung/hiclip}{[code]} & YFCC100M, CC3M, CC12M  & Con & Propose to incorporate hierarchy-aware attention into VLM pre-training.\\
        CLIPPO$\mathsection$~\cite{tschannen2022image}~\href{https://github.com/google-research/big_vision}{[code]} & WebLI$*$  & Con & Learn image and text data with a single network for VLM pre-training. \\ 
        OneR$\mathsection$~\cite{jang2023unifying} & CC3M, SBU, VG, COCO & Con,Gen & Unify image and text learning in a single tower transformer. \\

        RA-CLIP$\dagger$~\cite{xie2023ra} & YFCC100M &  Con &  Propose retrieval-augmented image-text contrastive learning.  \\
        LA-CLIP$\dagger$~\cite{fan2023improving}~\href{https://github.com/LijieFan/LaCLIP}{[code]} & CC3M, CC12M, RC, LAION400M &  {Con} &  {Propose LLMs-augmented image-text contrastive learning.}  \\
         {ALIP$\dagger$}~\cite{yang2023alip}~\href{https://github.com/deepglint/ALIP}{[code]} &  {YFCC100M} &  {Con} &  {Introduce synthetic caption supervision into VLM pre-training.} \\
         {GrowCLIP$\ddagger$}~\cite{deng2023growclip} &  {CC12M} &  {Con} &  {Propose online-learning image-text contrastive learning.}\\
        GroupVit$\dagger$~\cite{xu2022groupvit}~\href{https://github.com/NVlabs/GroupViT}{[code]} & CC12M, YMCC100M  & Con & Propose hierarchical visual concepts grouping for VLM pre-training. \\ 
        SegClip$\dagger$~\cite{luo2022segclip}~\href{https://github.com/ArrowLuo/SegCLIP}{[code]} & CC3M, COCO  & Con,Gen & Propose a plug-in semantic group module for VLM pre-training. \\ 
        CLIPpy$\dagger$~\cite{ranasinghe2022perceptual}~\href{https://github.com/ArrowLuo/SegCLIP}{[code]} & CC12M  & Con & Propose spatial representation aggregation for VLM pre-training. \\
        RegionClip$\dagger$~\cite{zhong2022regionclip}~\href{https://github.com/microsoft/RegionCLIP}{[code]} & CC3M, COCO  & Con,Align & Learn region-level visual representations for VLM pre-training.\\ 
        GLIP$\ddagger$~\cite{li2022grounded}~\href{https://github.com/microsoft/GLIP}{[code]} & CC3M, CC12M, SBU  & Align & 
        Unify detection and phrase grounding for grounded VLM pre-training.
        \\       FIBER$\ddagger$~\cite{doucoarse}~\href{https://github.com/microsoft/FIBER}{[code]} &  COCO, CC3M, SBU, VG  & Con,Gen,Align & Propose deep multi-modal fusion for coarse-to-fine VLM pre-training. \\
        DetCLIP$\ddagger$~\cite{yaodetclip} & YMCC100M & Align & Present a paralleled visual-concept VLM pre-training method. \\

    \bottomrule[1pt]
    \end{tabular}
    }
    \label{tab:sum_VLM}
\end{table*}

\section{Vision-Language Model Pre-training}\label{Sec.VLP}

VLM pre-training has been explored with three typical objectives: 
contrastive objectives, generative objectives and alignment objectives. This section reviews them with multiple VLM pre-training studies as listed in Table~\ref{tab:sum_VLM}.

\subsection{VLM Pre-Training with Contrastive Objectives}

Contrastive learning has been widely explored in VLM pre-training, which designs contrastive objectives for learning discriminative image-text features~\cite{mu2022slip,li2021supervision,radford2021learning}.

\subsubsection{Image Contrastive Learning}

This pre-training objective aims to learn discriminative features in image modality, which often serves as an auxiliary objective for fully exploiting the image data potential.
For example, SLIP~\cite{mu2022slip} employs a standard infoNCE loss defined in Eq.~\ref{eq:infonce} for learning discriminative image features.

\subsubsection{Image-Text Contrastive Learning}

\begin{figure}[ht]
    \centering
    \includegraphics[width=0.35\textwidth]{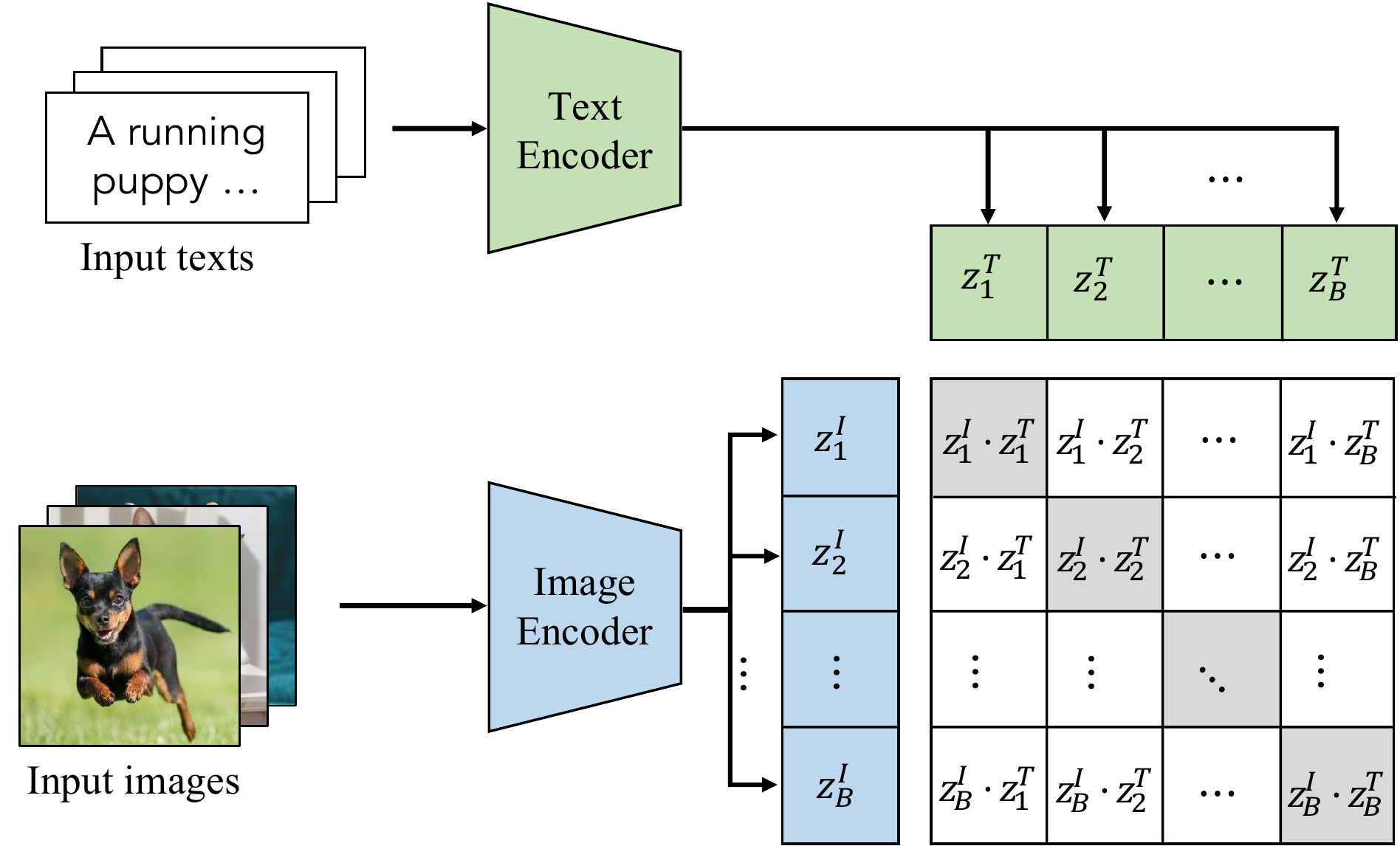}
    \caption{Illustration of the image-text contrastive learning in CLIP~\cite{radford2021learning}. 
    Figure is reproduced from~\cite{radford2021learning}.}
    \label{fig:clip}
\end{figure}

Image-text contrast aims to learn vision-language correlation by contrasting image-text pairs, $i.e.$, pulling the embeddings of paired images and texts close while pushing others faraway~\cite{radford2021learning}.  
For example, CLIP~\cite{radford2021learning} employs a symmetrical image-text infoNCE loss in Eq.~\ref{eq_img2text} which measures the image-text similarity by a dot-product between image and text embeddings in Fig.~\ref{fig:clip}. 
The pre-trained VLM hence learns image-text correlation which allows zero-shot predictions in downstream visual recognition tasks. 

Inspired by the great success of CLIP, many studies improve the symmetrical image-text infoNCE loss from different perspectives.
For example, ALIGN~\cite{jia2021scaling} scales up the VLM pre-training with large-scale ($i.e.$, 1.8 billions) but noisy image-text pairs with noise-robust contrastive learning. 
Several studies~\cite{li2021supervision,wu2021data,cui2022contrastive} instead explore data-efficient VLM pre-training with much less image-text pairs. 
For example, DeCLIP~\cite{li2021supervision} introduces nearest-neighbor supervision to utilize the information from similar pairs, enabling effective pre-training on limited data.
OTTER~\cite{wu2021data} employs optimal transport to pseudo-pair images and texts reducing the required training data greatly.
ZeroVL~\cite{cui2022contrastive} exploits limited data resource via debiased data sampling and data augmentation with coin flipping mixup.

Another line of follow-up studies~\cite{yao2021filip,gao2022pyramidclip,xu2022groupvit} aim for comprehensive vision-language correlation modelling by performing image-text contrastive learning across various semantic levels.
For example, FILIP~\cite{yao2021filip} introduces region-word alignment into contrastive learning, enabling to learn fine-grained vision-language corresponding knowledge.
PyramidCLIP~\cite{gao2022pyramidclip} constructs multiple semantic levels and performs both cross-level and peer-level contrastive learning for effective VLM pre-training.

Besides, several recent studies further improve by augmenting image-text pairs~\cite{fan2023improving,xie2023ra,yang2023alip,deng2023growclip}. For example, LA-CLIP~\cite{fan2023improving} and ALIP~\cite{yang2023alip} employ large language models to augment synthetic captions for given images while RA-CLIP~\cite{xie2023ra} retrieves relevant image-text pairs for image-text pair augmentation. To facilitate efficient communications across data modalities, \cite{jang2023unifying} and \cite{tschannen2022image} attempt to unify vision and language learning in a single encoder.

\subsubsection{Image-Text-Label Contrastive Learning}

\begin{figure}[ht]
    \centering
    \includegraphics[width=0.34\textwidth]{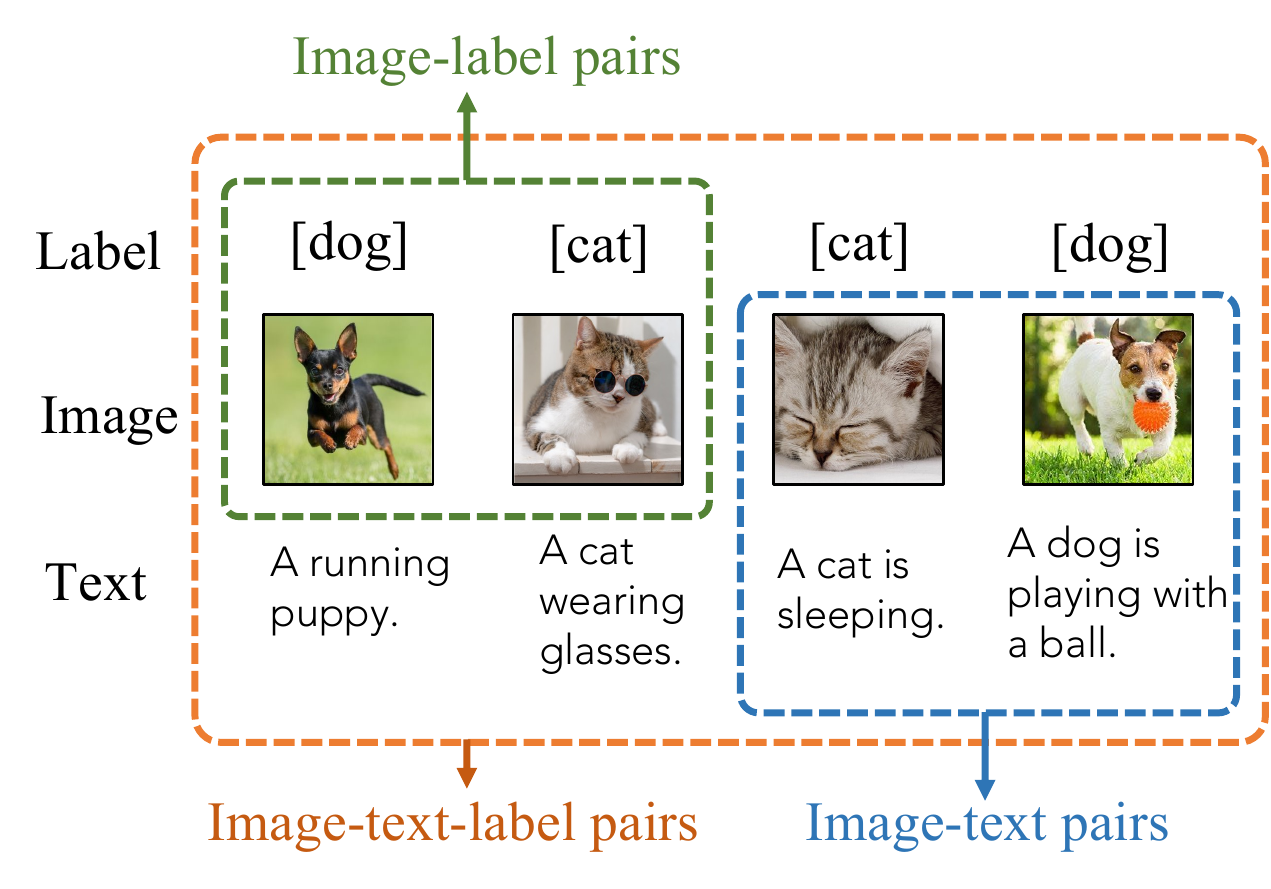}
    \caption{
    Illustration of the image-text-label space proposed in UniCL~\cite{yang2022unified}. Figure is reproduced from \cite{yang2022unified}.
    }
    \label{fig:UniCL}
\end{figure}

This type of pre-training introduces image classification labels~\cite{yang2022unified} into the image-text contrast as defined in Eq.~\ref{eq_img2text2label}, which encodes image, text and classification labels into a shared space as shown in Fig.~\ref{fig:UniCL}. It exploits both supervised pre-training with image labels and unsupervised VLM pre-training with image-text pairs. As reported in UniCL~\cite{yang2022unified}, such pre-training allows learning both discriminative and task-specific ($i.e.$, image classification) features simultaneously. The ensuing work in~\cite{yuan2021florence} scales UniCL with around 900M image-text pairs, leading to outstanding performance in various downstream recognition tasks.

\subsubsection{Discussion}

Contrastive objectives enforce positive pairs to have similar embeddings in contrast to negative pairs. They encourage VLMs to learn discriminative vision and language features~\cite{radford2021learning,jia2021scaling}, where more discriminative features generally lead to more confident and accurate zero-shot predictions.
However, the contrastive objective has two limitations: (1) Joint optimizing positive and negative pairs is complicated and challenging~\cite{radford2021learning,jia2021scaling}; (2) it involves a heuristic temperature hyper-parameter for controlling the feature discriminability as described in Sec.~\ref{sec.3.2.1}.

\subsection{VLM Pre-training with Generative Objectives}

Generative VLM pre-training learns semantic knowledge by learning to generate images or texts via masked image modelling, masked language modelling, masked cross-modal modelling and image-to-text generation.

\subsubsection{Masked Image Modelling}

\begin{figure}[ht]
    \centering
    \includegraphics[width=0.38\textwidth]{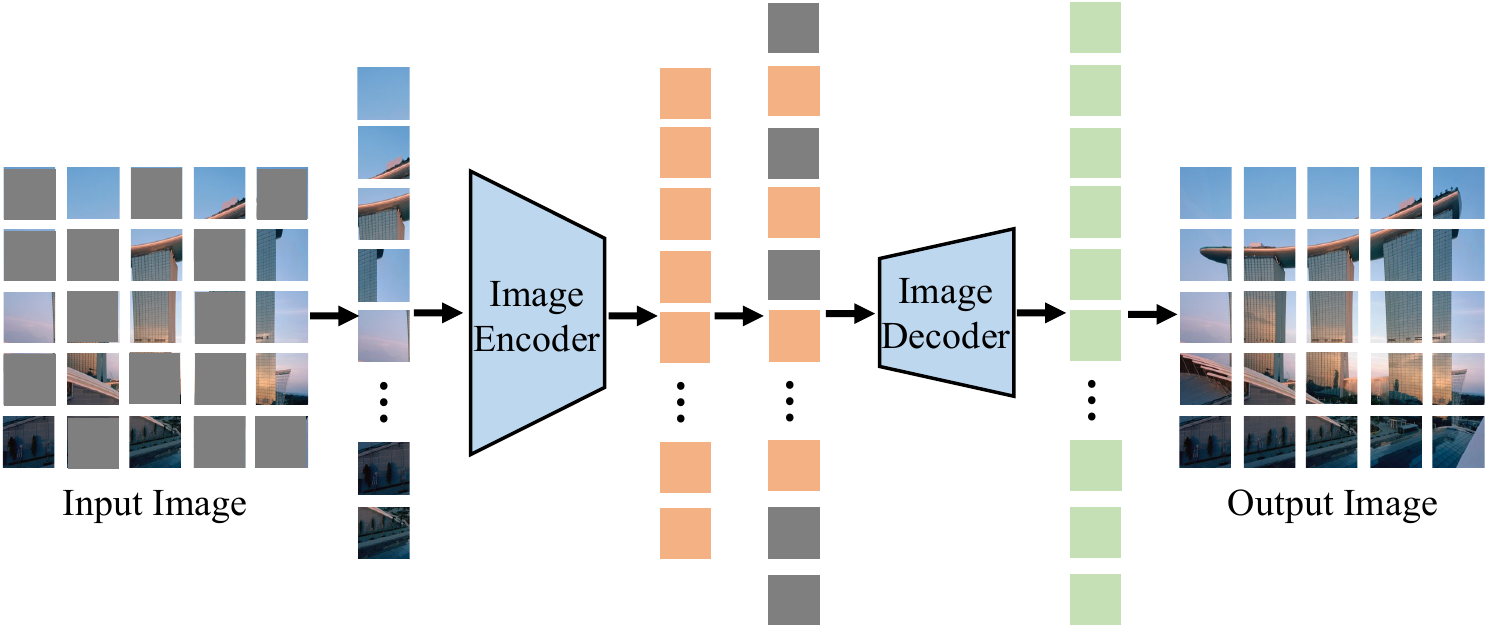}
    \caption{Illustration of masked image modelling~\cite{he2021masked}. 
    Figure is reproduced from~\cite{he2021masked}.}
    \label{fig:MIM}
\end{figure}

This pre-training objective guides to learn image context information by masking and reconstructing images as defined in Eq.~\ref{eq:MIM}.
In Masked Image Modelling ($e.g.$, MAE~\cite{he2022masked} and BeiT~\cite{bao2021beit}), certain patches in an image are masked and the encoder is trained to reconstruct them conditioned on unmasked patches as shown in Fig.~\ref{fig:MIM}. 
For example, FLAVA~\cite{singh2022flava} adopts rectangular block masking as in BeiT~\cite{bao2021beit}, while KELIP~\cite{ko2022large} and SegCLIP~\cite{luo2022segclip} follow MAE to mask out a large portion of patches ($i.e.$, 75 \%) in training.

\subsubsection{Masked Language Modelling}

\begin{figure}[ht]
    \centering
    \includegraphics[width=0.36\textwidth]{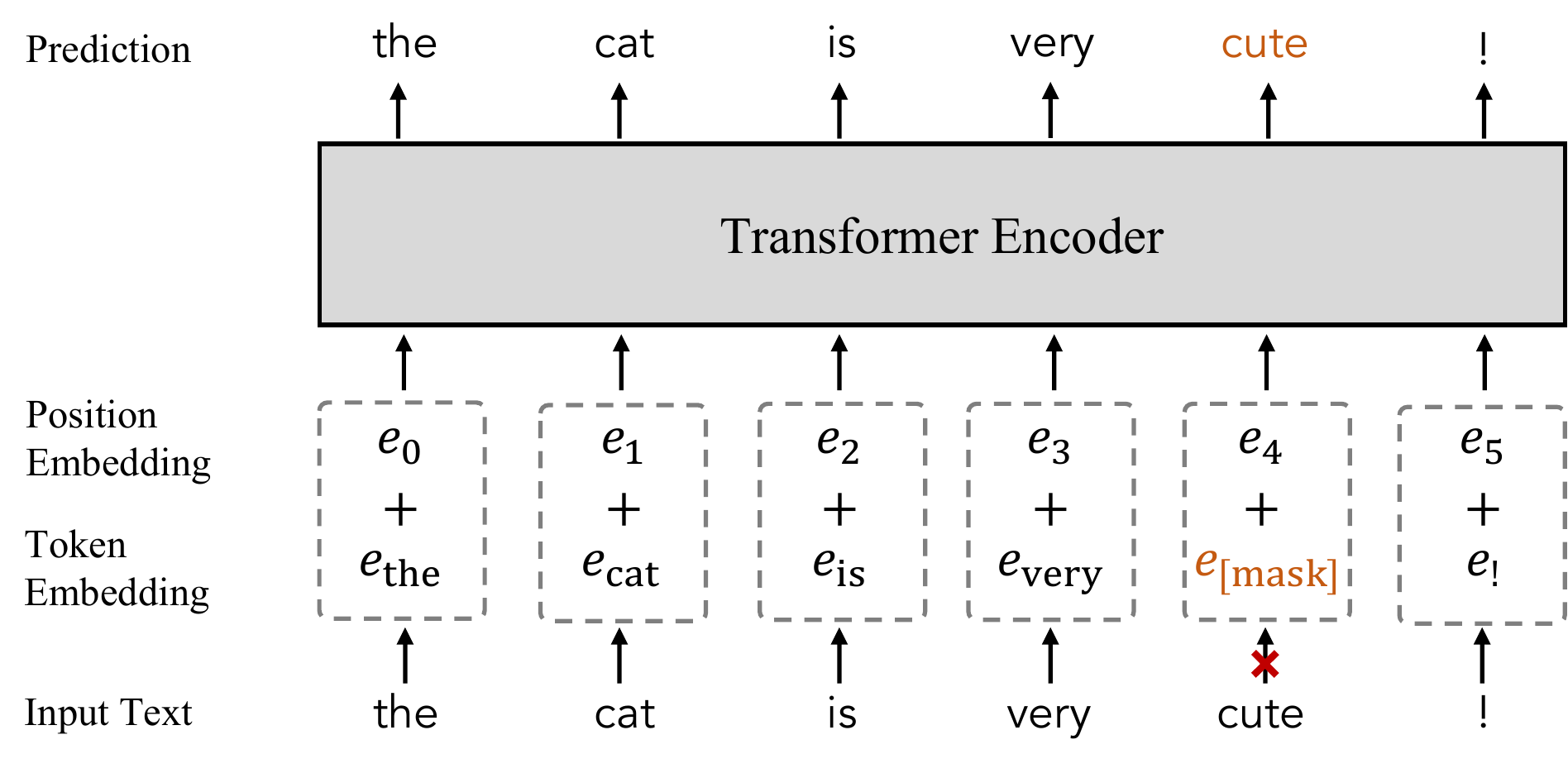}
    \caption{Illustration of masked language modelling~\cite{devlin2018bert}. 
    }
    \label{fig:MLM}
\end{figure}

Masked language modelling, a widely-adopted pre-training objective in NLP as defined in Eq.~\ref{eq:MLM}, also demonstrates its effectiveness in text feature learning in VLM pre-training.
It works by masking a fraction of tokens in each input text and training networks to predict the masked tokens as illustrated in Fig.~\ref{fig:MLM}. 
Following~\cite{devlin2018bert}, FLAVA~\cite{singh2022flava} masks out 15\% text tokens and reconstructs them from the rest tokens for modelling cross-word correlation.
FIBER~\cite{doucoarse} adopts masked language modelling~\cite{devlin2018bert} as one of the VLM pre-training objectives to extract better language features.

\subsubsection{Masked Cross-Modal Modelling}

Masked cross-modal modelling masks and reconstructs both image patches and text tokens jointly as defined in Eq.~\ref{eq:MMM}, which inherits the benefits of both masked image modelling and masked language modelling.
It works by masking a certain percentage of image patches and text tokens and training VLMs to reconstruct them based on the embeddings of unmasked image patches and text tokens.
For example, FLAVA~\cite{singh2022flava} masks $\sim$40\% image patches as in~\cite{bao2021beit} and 15\% text tokens as in~\cite{devlin2018bert}, and then employs a MLP to predict masked patches and tokens, capturing rich vision-language correspondence information.

\subsubsection{Image-to-Text Generation}

\begin{figure}[ht]
    \centering
    \includegraphics[width=0.35\textwidth]{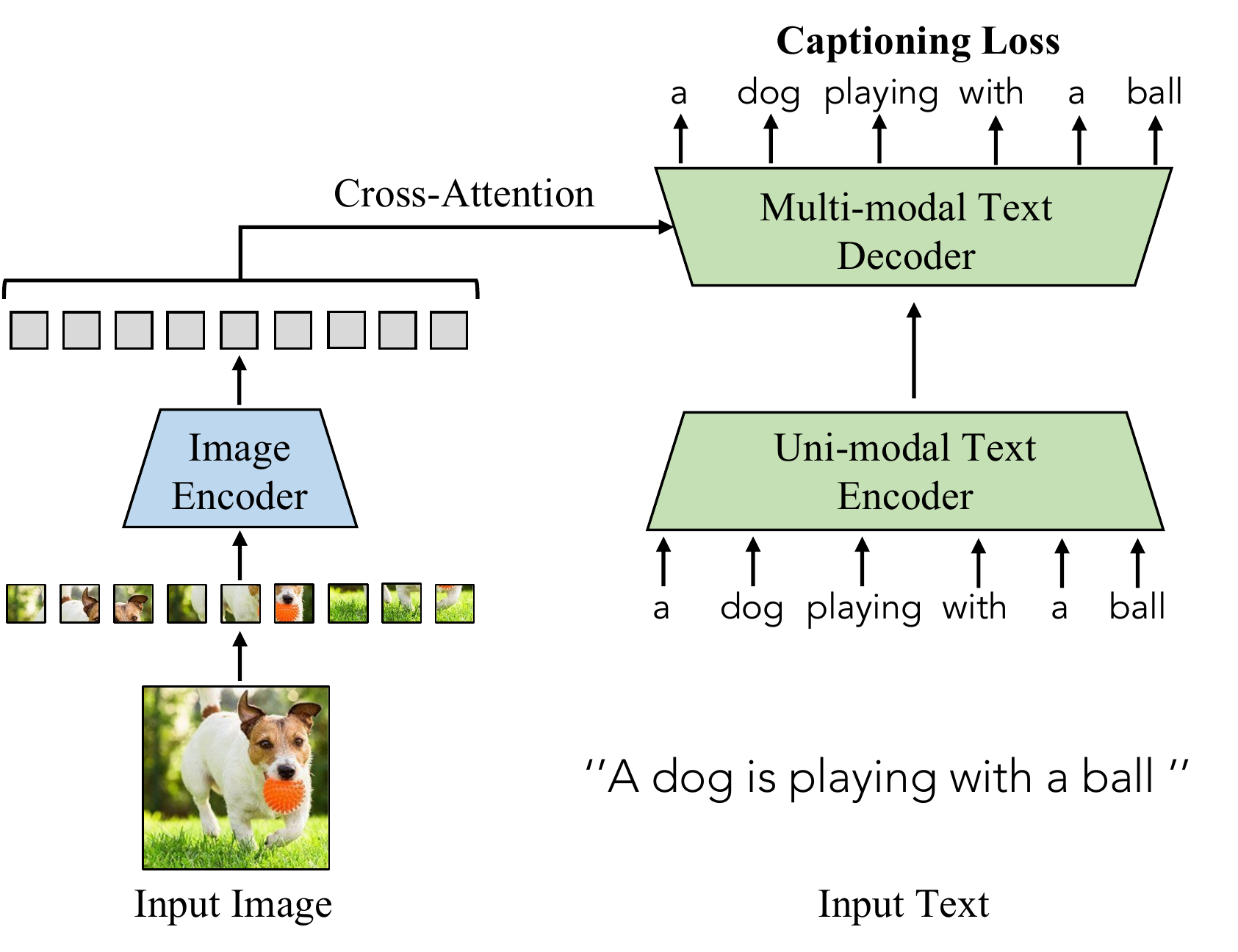}
    \caption{A simplified illustration of image-to-caption generation in COCA~\cite{yu2022coca}. Figure is reproduced based on~\cite{yu2022coca}.}
    \label{fig:coca}
\end{figure}

Image-to-text generation aims to generate descriptive texts for a given image for capturing fine-grained vision-language correlation by training VLMs to predict tokenized texts.
It first encodes an input image into intermediate embeddings and then decodes them into descriptive texts as defined in Eq.~\ref{eq:caption}.
For instance, COCA~\cite{yu2022coca}, NLIP~\cite{huang2022nlip} and PaLI~\cite{chen2022pali} train VLMs with the standard encoder-decoder architecture and image captioning objectives as shown in Fig.~\ref{fig:coca}.

\subsubsection{Discussion}

Generative objectives work by cross-modal generation or masked image/language/cross-modal modelling, encouraging VLMs to learn rich vision, language and vision-language contexts for better zero-shot predictions.
Hence, generative objectives are generally adopted as additional objectives above other VLM pre-training objectives for learning rich context information~\cite{yu2022coca,singh2022flava,li2021supervision}.

\subsection{VLM Pre-training with Alignment Objectives}
\label{sec.3.3}

Alignment objectives enforce VLMs to align paired images and texts  
by learning to predict whether the given text describes the given image correctly.
It can be broadly categorized into global image-text matching and local region-word matching for VLM pre-training. 

\subsubsection{Image-Text Matching}

Image-text matching models global image-text correlation by directly aligning paired images and texts as defined in Eq.~\ref{eq:ITM}.
For example, given a batch of image-text pairs, FLAVA~\cite{singh2022flava} matches the given image with its paired text via a classifier and a binary classification loss.
FIBER~\cite{doucoarse} follows~\cite{bao2021vlmo} to mine hard negatives with pair-wise similarities for better alignment between images and texts.

\begin{figure}[ht]
    \centering
    \includegraphics[width=0.36\textwidth]{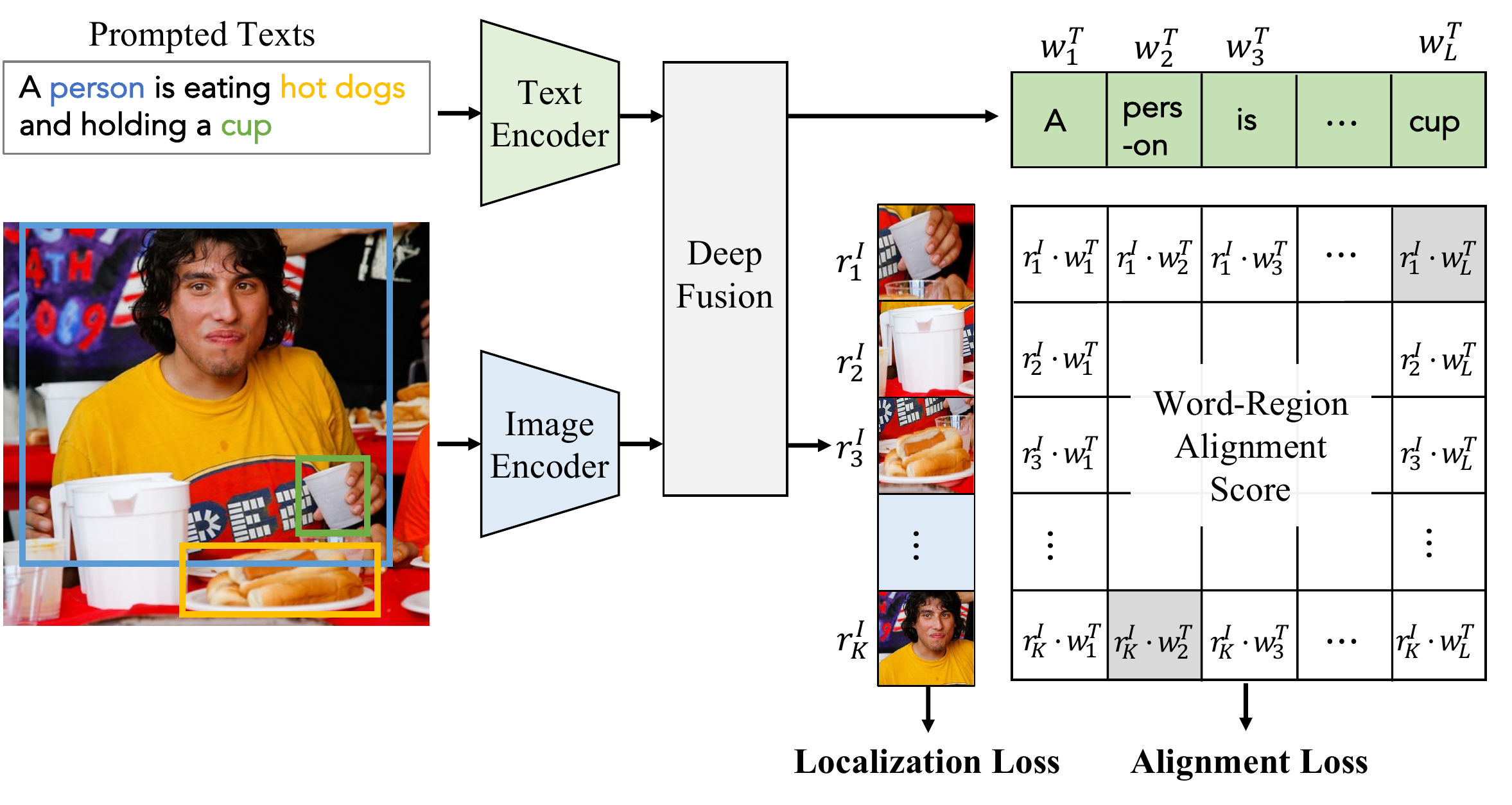}
    \caption{
    Illustration of GLIP~\cite{li2022grounded} that uses word-region alignment for detection. Figure is reproduced from \cite{li2022grounded}.
    }
    \label{fig:glip}
\end{figure}

\subsubsection{Region-Word Matching}

Region-word matching objective models local fine-grained vision-language correlation by aligning paired image regions and word tokens, greatly benefiting zero-shot dense predictions in object detection and semantic segmentation.
For example, GLIP~\cite{li2022grounded}, FIBER~\cite{doucoarse} and DetCLIP~\cite{yaodetclip} replace object classification logits by region-word alignment scores, $i.e.$, the dot-product similarity between regional visual features and token-wise features as illustrated in Fig.~\ref{fig:glip}.

\subsubsection{Discussion}

Alignment objectives learn to predict weather the given image and text data are matched or not, which are simple and easy-to-optimize and can be easily extended to model fine-grained vision-language correlation by matching image and text data locally.
On the other hand, they often learn little correlation information within vision or language modality.
Therefore, alignment objectives are often adopted as auxiliary losses to other VLM pre-training objectives for enhancing modelling the correlation across vision and language modalities~\cite{singh2022flava,zhou2022non}.

\subsection{Summary and Discussion}

In summary, VLM pre-training models the vision-language correlation with different cross-modal objectives such as image-text contrastive learning, masked cross-modal modelling, image-to-text generation and image-text/region-word matching.
Various single-modal objectives have also been explored for fully exploiting the data potential of its own modality, such as masked image modelling for image modality and masked language modelling for text modality. At the other end, recent VLM pre-training focuses on learning global vision-language correlation with benefits in image-level recognition tasks such as image classification. Meanwhile, several studies~\cite{xu2022groupvit,li2022grounded,luo2022segclip,zhong2022regionclip, yaodetclip,ranasinghe2022perceptual,doucoarse} model local fine-grained vision-language correlation via region-word matching, aiming for better dense predictions in object detection and semantic segmentation.

\begin{table*}[!t]
    \setlength\tabcolsep{6pt}
    \centering
    \caption{Summary of VLM transfer learning methods. TPT: text-prompt tuning; VPT: visual-prompt tuning; FA: feature adapter; CA: cross-attention; FT: fine-tuning; AM: architecture modification; LLM: large-language model. \textcolor{magenta}{[code]} directs to code websites.}
    \resizebox{0.98\linewidth}{!}{
    \begin{tabular}{l|c|c|p{11.5cm}}
    \toprule[1pt]
        \textbf{Method} & \textbf{Category} & \textbf{Setup} & \textbf{Contribution} \\
    \midrule
        CoOp~\cite{zhou2022learning}~\href{https://github.com/KaiyangZhou/CoOp}{[code]} & TPT & Few-shot Sup.  &  Introduce context optimization with learnable text prompts for VLM transfer learning.\\ 
        CoCoOp~\cite{zhou2022conditional}~\href{https://github.com/KaiyangZhou/CoOp}{[code]} & TPT & Few-shot Sup. & Propose conditional text prompting to mitigate overfitting in VLM transfer learning.\\ 
        SubPT~\cite{ma2022understanding}~\href{https://github.com/machengcheng2016/Subspace-Prompt-Learning}{[code]} & TPT & Few-shot Sup. & Propose subspace text prompt tuning to mitigate overfitting in VLM transfer learning.\\
        LASP~\cite{bulat2022language} & TPT & Few-shot Sup. & Propose to regularize the learnable text prompts with the hand-engineered prompts.\\
        ProDA~\cite{lu2022prompt} & TPT & Few-shot Sup. & Propose prompt distribution learning that captures the distribution of diverse text prompts. \\ 
        VPT~\cite{derakhshani2022variational} & TPT & Few-shot Sup. & Propose to model the text prompt learning with instance-specific distribution.\\
        ProGrad~\cite{zhu2022prompt}~\href{https://github.com/BeierZhu/Prompt-align}{[code]} & TPT & Few-shot Sup. & Present a prompt-aligned gradient technique for preventing knowledge forgetting.  \\
        CPL~\cite{he2022cpl}~\href{https://github.com/eric-ai-lab/CPL}{[code]} & TPT & Few-shot Sup. & Employ counterfactual generation and contrastive learning for text prompt tuning.\\
        PLOT~\cite{chen2022prompt}~\href{https://github.com/CHENGY12/PLOT}{[code]} & TPT & Few-shot Sup. & Introduce optimal transport to learn multiple comprehensive text prompts.\\
        DualCoOp~\cite{sundualcoop}~\href{https://github.com/sunxm2357/DualCoOp}{[code]} & TPT & Few-shot Sup. & Introduce positive and negative text prompt learning for multi-label classification. \\ 
        TaI-DPT~\cite{guo2022texts}~\href{https://github.com/guozix/TaI-DPT}{[code]} & TPT & Few-shot Sup. & Introduce a double-grained prompt tuning technique for multi-label classification\\
        SoftCPT~\cite{ding2022prompt}~\href{https://github.com/kding1225/softcpt}{[code]} & TPT & Few-shot Sup. & Propose to fine-tune VLMs on multiple downstream tasks simultaneously.\\
        DenseClip~\cite{rao2022denseclip}~\href{https://github.com/raoyongming/DenseCLIP}{[code]} & TPT &Supervised& Propose a language-guided fine-tuning technique for dense visual recognition tasks.\\ 
        UPL~\cite{huang2022unsupervised}~\href{https://github.com/tonyhuang2022/UPL}{[code]} & TPT & Unsupervised & Propose unsupervised prompt learning with self-training for VLM transfer learning.\\
        TPT~\cite{shutest}~\href{https://github.com/azshue/TPT}{[code]} & TPT &Unsupervised & Propose test-time prompt tuning that learns adaptive prompts on the fly.\\ 
         {KgCoOp}~\cite{yao2023visual}~\href{https://github.com/htyao89/KgCoOp}{[code]} &  {TPT} &  {Few-shot Sup.} &  {Introduce knowledge-guided prompt tuning to improve the generalization ability.}\\
         {ProTeCt}~\cite{wu2023protect} &  {TPT} &  {Few-shot Sup.} &   {Propose a prompt tuning technique to improve consistency of model predictions.}\\
        VP~\cite{bahng2022exploring}~\href{https://github.com/hjbahng/visual_prompting}{[code]} & VPT & Supervised & Investigate the efficacy of visual prompt tuning for VLM transfer learning. \\
         {RePrompt}~\cite{rong2023retrieval} &  {VPT} &  {Few-shot Sup.} &  {Introduce retrieval mechanisms to leverage knowledge from downstream tasks.}\\
        UPT~\cite{zang2022unified}~\href{https://github.com/yuhangzang/upt}{[code]} & TPT, VPT & Few-shot Sup. & Propose a unified prompt tuning that jointly optimizes text and image prompts.\\
        
        MVLPT~\cite{shen2022multitask}\href{https://github.com/facebookresearch/vilbert-multi-task}{[code]} & TPT, VPT & Few-shot Sup. & Incorporate multi-task knowledge into text and image prompt tuning. \\
        MaPLE~\cite{khattak2022maple}\href{https://github.com/muzairkhattak/multimodal-prompt-learning}{[code]} & TPT, VPT & Few-shot Sup. & Propose multi-modal prompt tuning with a mutual promotion strategy.\\
        CAVPT~\cite{xing2022class}\href{https://github.com/fanrena/DPT}{[code]} & TPT, VPT & Few-shot Sup. & Introduce class-aware visual prompt for concentrating more on visual concepts.\\
        Clip-Adapter~\cite{gao2021clip}\href{https://github.com/gaopengcuhk/CLIP-Adapter}{[code]} & FA & Few-shot Sup. & Introduce an adapter with residual feature blending for efficient VLM transfer learning. \\
        Tip-Adapter~\cite{zhang2021tip}\href{https://github.com/gaopengcuhk/Tip-Adapter}{[code]} & FA & Few-shot Sup. & 
        Propose to build a training-free adapter with the embeddings of few labelled images.
        \\
        SVL-Adpter~\cite{pantazis2022svl}\href{https://github.com/omipan/svl_adapter}{[code]} & FA & Few-shot Sup.& Introduce a self-supervised adapter by performing self-supervised learning on images. \\
        SuS-X~\cite{udandarao2022sus}\href{https://github.com/vishaal27/sus-x}{[code]} & FA & Unsupervised & Propose a training-free name-only transfer learning paradigm with curated support sets.\\
        CLIPPR~\cite{kahana2022improving}\href{https://github.com/jonkahana/CLIPPR}{[code]} & FA & Unsupervised & Leverage the label distribution priors for adapting pre-trained VLMs. \\
        SgVA-CLIP~\cite{peng2022sgva} & TPT, FA & Few-shot Sup. & Propose a semantic-guided visual adapter to generate discriminative adapted features.\\ 
        VT-Clip~\cite{zhang2021vt} & CA & Few-shot Sup. & 
        Introduce visual-guided attention that semantically aligns text and image features.
        \\ 
        CALIP~\cite{guo2022calip}~\href{https://github.com/ziyuguo99/calip}{[code]} & CA & Unsupervised & Propose parameter-free attention for the communication between visual and textual features.\\
         {TaskRes}~\cite{yu2023task}~\href{https://github.com/geekyutao/TaskRes}{[code]} &  {CA} &  {Few-shot Sup.} &  {Propose a technique for better learning old VLM knowledge and new task knowledge.} \\
        {CuPL}~\cite{pratt2022does} & {LLM} & {Unsupervised} &{Employ large language models to generate customized prompts for VLMs. }\\
         {VCD}~\cite{menon2022visual} &  {LLM} &  {Unsupervised} &  {Employ large language models to generate captions for VLMs.}\\
        Wise-FT~\cite{wortsman2022robust}\href{https://github.com/mlfoundations/wise-ft}{[code]} & FT &Supervised& Propose ensemble-based fine-tuning by combining the fine-tuned and original VLMs. 
        \\ 
        MaskClip~\cite{zhou2022extract}\href{https://github.com/chongzhou96/MaskCLIP}{[code]} & AM & Unsupervised & Propose to extract dense features by modifying the image encoder architecture. \\ 
        MUST~\cite{li2022masked}\href{https://github.com/salesforce/MUST}{[code]} & Self-training & Unsupervised & Propose masked unsupervised self-training for unsupervised VLM transfer learning.\\
    \bottomrule[1pt]
    \end{tabular}
    }
    \label{tab:sum_TL}
\end{table*}

\section{VLM Transfer Learning}
\label{sec:TL}

Beyond \textit{zero-shot prediction} that directly applies pre-trained VLMs on downstream tasks without fine-tuning, transfer learning has been studied recently which adapts VLMs to fit downstream tasks via prompt tuning~\cite{zhou2022learning,ma2022understanding}, feature adapter~\cite{gao2021clip,zhang2021tip}, etc. This section presents the motivation of transfer learning for pre-trained VLMs, the common transfer-learning setup, and three transfer learning approaches including prompt tuning methods, feature adapter methods and other methods.

\subsection{Motivation of Transfer learning}
\label{sec:moti_transfer}

Although pre-trained VLMs have demonstrated strong generalization capability, they often face two types of gaps while applied to various downstream tasks: 1) the gaps in image and text distributions, $e.g.$, an downstream dataset may have task-specific image styles and text formats; 2) the gaps in training objectives, $e.g.$, VLMs are generally trained with task-agnostic objectives and learn general concepts while downstream tasks often involve task-specific objectives such as coarse or fine-grained classification, region or pixel-level recognition, etc.

\subsection{Common Setup of Transfer Learning}

Three transfer setups have been explored for mitigating the domain gaps described in Sec.~\ref{sec:moti_transfer}, including supervised transfer, few-shot supervised transfer and unsupervised transfer. Supervised transfer employs all labelled downstream data for fine-tuning the pre-trained VLMs, while few-shot supervised transfer is more annotation efficient which just uses a small amount of labelled downstream samples. Differently, unsupervised transfer uses unlabelled downstream data for fine-tuning VLMs. It is thus more challenging but more promising and efficient for VLM transfer.

\subsection{Common Transfer Learning Methods
}

As shown in Table~\ref{tab:sum_TL}, 
we broadly group existing VLM transfer methods into three categories including prompt tuning approaches, feature adapter approaches, and others.

\subsubsection{Transfer via Prompt Tuning}

Inspired by the ``prompt learning" in NLP~\cite{liu2023pre}, many VLM prompt learning methods have been proposed for adapting VLMs to fit downstream tasks by finding optimal prompts without fine-tuning the entire VLM.
Most existing studies follow three approaches by text prompt tuning, visual prompt tuning, and text-visual prompt tuning.

\begin{figure}[ht]
    \centering
    \includegraphics[width=0.44\textwidth]{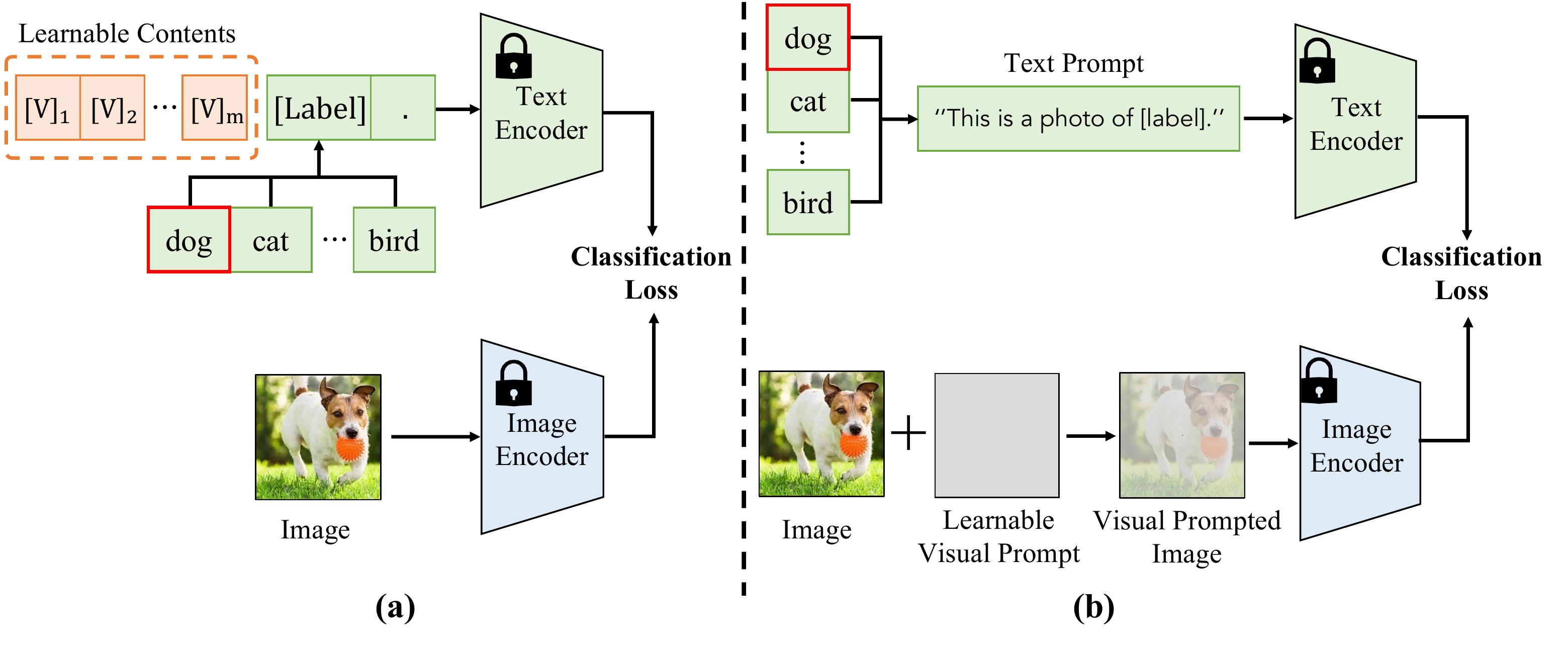}
    \caption{
    Illustration of text prompt learning~\cite{zhou2022learning} in (a) and visual prompt learning~\cite{bahng2022exploring} in (b). 
    }
    \label{fig:coop}
\end{figure}

\noindent \textbf{Transfer with Text Prompt Tuning.} 
Different from prompt engineering~\cite{liu2023pre} that manually designs text prompts for each task, text prompt tuning explores more effective and efficient learnable text prompts with several labelled downstream samples for each class.
For example, CoOp~\cite{zhou2022learning} explores context optimization to learn context words for a single class name with learnable word vectors.  
It expands a category word [\verb|label|] into a sentence `[V]$_1$, [V]$_2$, ..., [V]$_m$ [\verb|label|]', where [V] denotes the learnable word vectors that are optimized by minimizing the classification loss with the downstream samples as shown in Fig.~\ref{fig:coop} (a). 
To mitigate the overfitting due to limited downstream samples in prompt learning, CoCoOp~\cite{zhou2022conditional} explores conditional context optimization that generates a specific prompt for each image.
SubPT~\cite{ma2022understanding} designs subsapce prompt tuning to improve the generalization of learned prompts. 
LASP~\cite{bulat2022language} regularizes learnable prompts with hand-engineered prompts.
VPT~\cite{derakhshani2022variational} models text prompts with instance-specific distribution with better generalization on downstream tasks.
KgCoOp~\cite{yao2023visual} enhances the generalization of unseen class by mitigating the forgetting of textual knowledge.

In addition, SoftCPT~\cite{ding2022prompt} fine-tunes VLMs on multiple few-shot tasks simultaneously for benefiting from multi-task learning.
PLOT~\cite{chen2022prompt} employs optimal transport to learn multiple prompts to describe the diverse characteristics of a category.
DualCoOp~\cite{sundualcoop} and TaI-DP~\cite{guo2022texts} transfer VLMs to multi-label classification tasks, where DualCoOp adopts both positive and negative prompts for multi-label classification while TaI-DP introduces double-grained prompt tuning for capturing both coarse-grained and fine-grained embeddings. 
DenseCLIP~\cite{rao2022denseclip} explores language-guided fine-tuning that employs visual features to tune text prompts for dense prediction~\cite{ren2015faster,chen2017deeplab}.
ProTeCt~\cite{wu2023protect} improves the consistency of model predictions for hierarchical classification task.

Beyond supervised and few-shot supervised prompt learning, recent studies explore unsupervised prompt tuning for better annotation efficiency and scalability. 
For instance, UPL~\cite{huang2022unsupervised} optimizes learnable prompts with self-training on selected pseudo-labeled samples.
TPT~\cite{shutest} explores test-time prompt tuning to learn adaptive prompts from a single downstream sample. 

\noindent \textbf{Transfer with Visual Prompt Tuning.}
Unlike text prompt tuning, visual prompt tuning~\cite{jia2022visual,rong2023retrieval} transfers VLMs by modulating the input of image encoder as shown in Fig.~\ref{fig:coop} (b).
For example, VP~\cite{bahng2022exploring} adopts learnable image perturbations $v$ to modify the input image $x^I$ by $x^I + v$, aiming to adjust $v$ to minimize a recognition loss. %
RePrompt~\cite{rong2023retrieval} integrates retrieval mechanisms into visual prompt tuning, allowing leveraging the knowledge from downstream tasks.
Visual prompt tuning enables pixel-level adaptation to downstream tasks, benefiting them greatly especially for dense prediction tasks.

\noindent \textbf{Transfer with Text-Visual Prompt Tuning} aims to modulate the text and image inputs simultaneously, benefiting from joint prompt optimization on multiple modalities.
For example, UPT~\cite{zang2022unified} unifies prompt tuning to jointly optimize text and image prompts, demonstrating the complementary nature of the two prompt tuning tasks. 
MVLPT~\cite{shen2022multitask} explores multi-task vision-language prompt tuning to incorporate cross-task knowledge into text and image prompt tuning.
MAPLE~\cite{khattak2022maple} conducts multi-modal prompt tuning by aligning visual prompts with their corresponding language prompts, enabling a mutual promotion between text prompts and image prompts.
CAVPT~\cite{xing2022class} introduces a cross attention between class-aware visual prompts and text prompts, encouraging the visual prompts to concentrate more on visual concepts.

\begin{figure}[ht]
    \centering
    \includegraphics[width=0.38\textwidth]{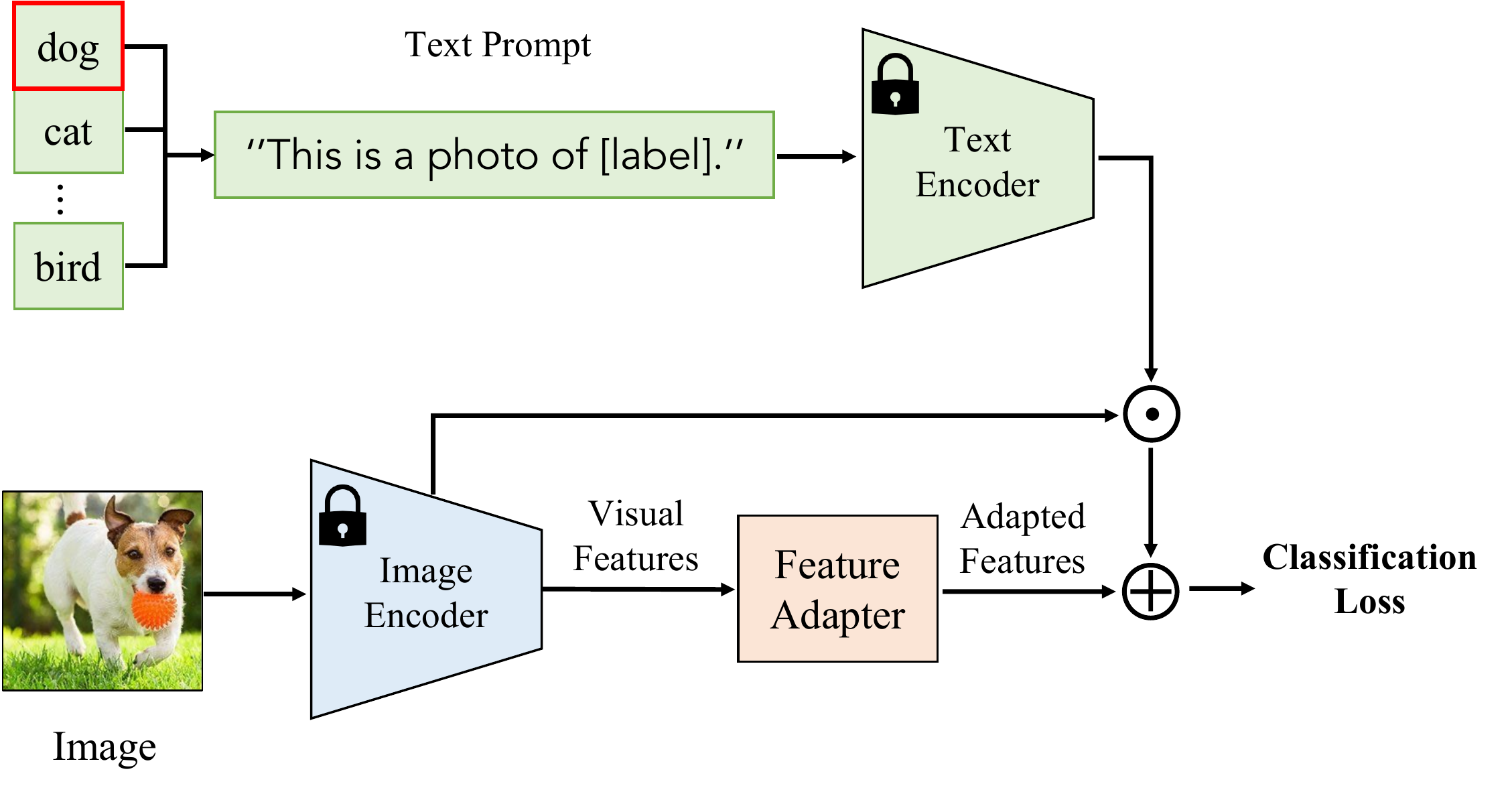}
    \caption{Illustration of feature adapter~\cite{gao2021clip}. 
    }
    \label{fig:adapter}
\end{figure}

\noindent
\textbf{Discussion.}
Prompt tuning enables parameter-efficient VLM transfer by modifying input texts/images with a few learnable text/image prompts. 
It is simple and easy-to-implement, and requires little extra network layers or complex network modifications.
Therefore, prompt tuning allows adapting VLMs in a black-box manner, which has clear advantages in transferring VLMs that involve concerns in intellectual property.
However, it still suffers from several limitations such as the low flexibility by following the manifold of the original VLMs in prompting~\cite{zhou2022learning}.

\subsubsection{Transfer via Feature Adaptation}

Feature adaptation fine-tunes VLMs to adapt image or text features with an additional light-weight feature adapter~\cite{houlsby2019parameter}. For example, Clip-Adapter~\cite{gao2021clip} inserts several trainable linear layers after CLIP's language and image encoders and optimizes them while keeping CLIP architecture and parameters frozen 
as illustrated in Fig.~\ref{fig:adapter}. Tip-Adapter~\cite{zhang2021tip} presents a training-free adapter that directly employs the embeddings of few-shot labelled images as the adapter weights. SVL-Adapter~\cite{pantazis2022svl} designs a self-supervised adapter which employs an additional encoder for self-supervised learning on input images. In summary, feature adapter adapts image and text features to fit VLMs to downstream data, which provides a promising alternative to prompt tuning for VLMs transfer.

\noindent {\textbf{Discussion.}
Feature adaptation adapts VLMs by modifying image and text features with an additional light-weight feature adapter. It is flexible and effective as its architecture and the insertion manner allow tailoring flexibly for different downstream tasks.
Therefore, feature adaptation has clear advantages in adapting VLMs to work on very different and complex downstream tasks~\cite{zhang2022pointclip,huang2023clip2point,xu2023side,chen2023tem}.
On the other hand, it requires modifying network architecture and thus can not handle VLMs that have concerns in intellectual property.
}

\subsubsection{Other Transfer Methods}

Several studies transfer VLMs by direct fine-tuning~\cite{wortsman2022robust}, architecture modification~\cite{zhou2022extract}, and cross attention~\cite{zhang2021vt,guo2022calip}. Specifically, Wise-FT~\cite{wortsman2022robust} combines the weights of a fine-tuned VLM and the original VLM for learning new information from downstream tasks.
MaskCLIP~\cite{zhou2022extract} extracts dense image features by modifying the architecture of the CLIP image encoder. 
VT-CLIP~\cite{zhang2021vt} introduces visual-guided attention to semantically correlate text features with downstream images, leading to a better transfer performance. CALIP~\cite{guo2022calip} introduces parameter-free attention for effective interaction and communication between visual and text features, leading to text-aware image features and visual-guided text features. 
TaskRes~\cite{yu2023task} directly tunes text-based classifier to exploit the old knowledge in the pre-trained VLM.
CuPL~\cite{pratt2022does} and VCD~\cite{menon2022visual} employ large language models, $e.g.$, GPT$_3$~\cite{brown2020language}, to augment text prompts for learning rich discriminative text information.

\subsection{Summary and Discussion}

In summary, prompt tuning and feature adapter are two major approaches for VLM transfer which work by modifying the input text/image and adapting image/text features, respectively. In addition, both approaches introduce very limited parameters while freezing the original VLMs, leading to efficient transfer. Further, while most studies follow few-shot supervised transfer~\cite{zhou2022learning,zhou2022conditional, ma2022understanding, lu2022prompt}, recent studies show that unsupervised VLM transfer can achieve competitive performance on various tasks~\cite{pratt2022does, huang2022unsupervised,shutest}, inspiring more research on unsupervised VLM transfer.

\section{VLM Knowledge Distillation}\label{Sec.KD}

As VLMs capture generalizable knowledge that covers a wide range of visual and text concepts, several studies explore how to distil the general and robust VLM knowledge while tackling complex dense prediction tasks such as object detection and semantic segmentation. This section presents the motivation of distilling knowledge from VLMs as well as two groups of knowledge distillation studies on the tasks of semantic segmentation and object detection. 

\subsection{Motivation of Distilling Knowledge from VLMs}

Different from VLM transfer that generally keeps the original VLM architecture intact in transfer~\cite{zhou2022learning,ma2022understanding,zhu2022prompt}, VLM knowledge distillation distils general and robust VLM knowledge to task-specific models without the restriction of VLM architecture, benefiting task-specific designs while tackling various dense prediction tasks~\cite{zang2022open, gu2021open,zhou2022zegclip}. For example, knowledge distillation allows transferring the general VLM knowledge to tackle detection tasks while taking the advantages of state-of-the-art detection architectures such as Faster R-CNN~\cite{ren2015faster} and DETR~\cite{carion2020end}.

\subsection{Common Knowledge Distillation Methods}

As VLMs are generally pre-trained with architectures and objectives designed for image-level representation, most VLM knowledge distillation methods focus on transferring image-level knowledge to region- or pixel-level tasks such as object detection and semantic segmentation. 
Table~\ref{tab. Sum_KD} shows a list of VLM knowledge distillation methods.

\begin{table*}[!ht]
    \centering
    \caption{Summary of VLM knowledge distillation methods. \textcolor{magenta}{[code]} directs to code websites.}
    \resizebox{0.98\linewidth}{!}{
    \begin{tabular}{p{1.6cm}|l|p{12.cm}}
    \toprule
    \textbf{Task} & \textbf{Method} &\textbf{ Contribution} \\
    \midrule
    Semantic &CLIPSeg~\cite{luddecke2022image}~\href{https://github.com/timojl/clipseg}{[code]}  & Extend CLIP by introducing a lightweight transformer-based decoder.\\ 
    Segmentation&ZegFormer~\cite{ding2022decoupling}~\href{https://github.com/dingjiansw101/ZegFormer}{[code]} & Group the pixels into segments and preforms zero-shot classification task on the segments. \\ 
    &LSeg~\cite{li2021language}~\href{https://github.com/isl-org/lang-seg}{[code]} &  
    Propose language-driven semantic segmentation by matching pixel and text embeddings.
    \\ 
    &SSIW~\cite{zabari2021semantic} & 
    Introduce a test-time augmentation technique to refine the pseudo labels generated by CLIP.
    \\
    &MaskClip+~\cite{zhou2022extract}~\href{https://github.com/chongzhou96/MaskCLIP}{[code]}  & Perform self-training with the pseudo labels generated by MaskClip (modified from CLIP).
    \\ 
    &ZegClip~\cite{zhou2022zegclip}~\href{https://github.com/ZiqinZhou66/ZegCLIP}{[code]} & Propose deep prompt tuning, non-mutually exclusive loss and relationship descriptor. \\
    &Fusioner~\cite{ma2022fusion}~\href{https://github.com/chaofanma/Fusioner}{[code]}  & Introduce cross-modality fusion that aligns the visual representation with language concept. \\ %
    &OVSeg~\cite{liang2022open}~\href{https://github.com/facebookresearch/ov-seg}{[code]}  & Adapt CLIP with the region-word pairs generated by the modified MaskFormer.\\
    &ZSSeg~\cite{xu2022simple}~\href{https://github.com/MendelXu/zsseg.baseline}{[code]} & Propose to first generate mask proposals and then classifies the generated mask proposals.\\ 
    &OpenSeg~\cite{ghiasi2022scaling}~\href{https://github.com/tensorflow/tpu/tree/641c1ac6e26ed788327b973582cbfa297d7d31e7/models/official/detection/projects/openseg}{[code]}  & Propose to align each word in the caption with the generated segmentation masks.\\ 
    &ReCo~\cite{shinreco}~\href{https://github.com/NoelShin/reco}{[code]}  & 
    Propose language-guided co-segmentation with the CLIP-retrieved images. 
    \\ 
    &CLIMS~\cite{xie2022clims} ~\href{https://github.com/CVI-SZU/CLIMS}{[code]}  & 
    Use CLIP to generate high-quality class activation maps w/o involving irrelevant background.
    \\ 
    &CLIP-ES~\cite{lin2022clip} ~\href{https://github.com/linyq2117/CLIP-ES}{[code]}  & 
    Employ CLIP to refine the class activation map for weakly-supervised segmentation.
    \\
    &FreeSeg~\cite{qin2023freeseg}~\href{https://freeseg.github.io/}{[code]} & Propose a unified, universal and open-Vocabulary image segmentation network.\\
    \midrule
    Object& ViLD~\cite{gu2021open}~\href{https://github.com/tensorflow/tpu/tree/master/models/official/detection/projects/vild}{[code]} & Propose to distill knowledge from a pre-trained VLM into a two-stage object detector.\\ 
    Detection&DetPro~\cite{du2022learning}~\href{https://github.com/dyabel/detpro}{[code]}  &  Propose to learn continuous prompt representations for open-vocabulary object detection.\\
    &HierKD~\cite{ma2022open}~\href{https://github.com/mengqiDyangge/HierKD}{[code]}  &Propose hierarchical knowledge distillation for global-level and instance-level distillation.\\ 
    &RKD~\cite{rasheedbridging}~\href{https://github.com/hanoonaR/object-centric-ovd}{[code]}   & Propose region-based knowledge distillation for aligning region- and image-level embeddings. \\ 
    &PromptDet~\cite{feng2022promptdet}~\href{https://github.com/fcjian/PromptDet}{[code]} & Introduce regional prompting for aligning text embeddings with regional image embeddings. \\ 
    &PB-OVD~\cite{gao2022open}~\href{https://github.com/salesforce/PB-OVD}{[code]}  & Propose to train object detectors with the pseudo bounding-box labels generated by VLMs. \\
    &CondHead~\cite{wang2022learning}  & Propose semantic-visual alignment for better box regression and mask segmentation.  \\
    &VLDet~\cite{lin2022learning}~\href{https://github.com/clin1223/VLDet}{[code]}  & 
    Achieve open-vocabulary object detection by the bipartite matching between regions and words.
    \\
    &F-VLM~\cite{kuo2022f}  & Propose to simply build a detection head upon the pre-trained VLM for object localization.\\ 
    &OV-DETR~\cite{zang2022open}~\href{https://github.com/yuhangzang/OV-DETR}{[code]}  & Achieve open-vocabulary detection transformer with a binary matching strategy.\\ 
    &Detic~\cite{zhou2022detecting}~\href{https://github.com/facebookresearch/Detic}{[code]}  & 
    Enlarge detection vocabulary using image-level supervision and pre-trained CLIP text encoder.
    \\ 
    &XPM~\cite{huynh2022open}~\href{https://github.com/hbdat/cvpr22_cross_modal_pseudo_labeling}{[code]}  & Design cross-modal pseudo-labeling to let VLMs generate caption-driven pseudo masks. 
    \\
    &OWL-ViT~\cite{minderer2022simple}~\href{https://github.com/google-research/scenic/tree/main/scenic/projects/owl_vit}{[code]}  & 
    Propose ViT-based open-vocabulary detector by adding object classification/localization head.
    \\ 
    &VL-PLM~\cite{zhao2022exploiting}~\href{https://github.com/xiaofeng94/VL-PLM}{[code]}  & Leverage VLMs for assigning category labels to the generated pseudo bounding boxes. \\ 
    &P$^3$OVD~\cite{long2022p}  & Propose prompt-driven self-training that refines the pseudo labels generated by VLMs. \\

    &ZSD-YOLO~\cite{xie2021zsd}~\href{https://github.com/Johnathan-Xie/ZSD-YOLO}{[code]}  & Leverage CLIP for object detection with a self-labeling based data augmentation techiqniue. \\
    &RO-ViT~\cite{kim2023region} & Bridge the gap of VLM pre-training and downstream open-vocabulary detection.\\
    &BARON~\cite{wu2023aligning}~\href{https://github.com/wusize/ovdet}{[code]} & Propose neighborhood sampling strategy to align the embedding of bag of regions. \\
    &OADP~\cite{wang2023object}~\href{https://github.com/LutingWang/OADP}{[code]} & Propose object-aware distillation network to preserve and transfer contextual knowledge. \\
    \bottomrule
    \end{tabular}
    }
    \label{tab. Sum_KD}
\end{table*}

\subsubsection{Knowledge Distillation for Object Detection}

Open-vocabulary object detection~\cite{zhou2022detecting} aims to detect objects described by arbitrary texts, $i.e.$, objects of any categories beyond the base classes. As VLMs like CLIP are trained with billion-scale image-text pairs that cover very broad vocabulary, many studies explore to distill VLM knowledge to enlarge the detector vocabulary. For example, ViLD~\cite{gu2021open} distills VLM knowledge to a two-stage detector whose embedding space is enforced to be consistent with that of CLIP image encoder. Following ViLD, HierKD~\cite{ma2022open} explores hierarchical global-local knowledge distillation, and RKD~\cite{rasheedbridging} explores region-based knowledge distillation for better aligning region-level and image-level embeddings. ZSD-YOLO~\cite{xie2021zsd} introduces self-labelling data augmentation for exploiting CLIP for better object detection. 
OADP~\cite{wang2023object} preserves proposal features while transferring contextual knowledge.
BARON~\cite{wu2023aligning} uses neighborhood sampling to distill a bag of regions instead of individual regions.
RO-ViT~\cite{kim2023region} distills regional information from VLMs for open-vocabulary detection.

Another line of research explores VLM distillation via prompt learning~\cite{liu2023pre}. For example, DetPro~\cite{du2022learning} introduces a detection prompt technique for learning continuous prompt representations for open-vocabulary object detection. PromptDet~\cite{feng2022promptdet} introduces regional prompt learning for aligning word embeddings with regional image embeddings. Additionally, several studies~\cite{gao2022open, huynh2022open,long2022p,ghiasi2022scaling,xu2022simple} explore VLM-predicted pseudo labels to improve object detectors. For example, PB-OVD~\cite{gao2022open} trains object detectors with VLM-predicted pseudo bounding boxes while XPM~\cite{huynh2022open} introduces a robust cross-modal pseudo-labeling strategy that employs VLM-generated pseudo masks for open-vocabulary instance segmentation. P$^3$OVD~\cite{long2022p} exploits prompt-driven self-training that refines the VLM-generated pseudo labels with fine-grained prompt tuning.

\subsubsection{Knowledge Distillation for Semantic Segmentation}

\textbf{Knowledge distillation for open-vocabulary semantic segmentation} leverages VLMs to enlarge the vocabulary of segmentation models, aim to segment pixels described by arbitrary texts ($i.e.$, any categories of pixels beyond base classes). For example, \cite{ding2022decoupling,ghiasi2022scaling,xu2022simple} achieve open-vocabulary semantic segmentation by first class-agnostic segmentation by grouping pixels into multiple segments and then segment recognition  with CLIP. CLIPSeg~\cite{luddecke2022image} introduces a lightweight transformer decoder to extend CLIP for semantic segmentation. LSeg~\cite{li2021language} maximizes the correlation between CLIP text embeddings and pixel-wise image embedding encoded by segmentation models.  
ZegCLIP~\cite{zhou2022zegclip} employs CLIP to generate semantic masks and introduces a relationship descriptor to mitigate overfitting on base classes. 
MaskCLIP+~\cite{zhou2022extract} and SSIW~\cite{zabari2021semantic} distill knowledge with VLM-predicted pixel-level pseudo labels. 
FreeSeg~\cite{qin2023freeseg} generates mask proposals firstly and then performs zero-shot classification for them.

\noindent\textbf{Knowledge distillation for weakly-supervised semantic segmentation} aims to leverage both VLMs and weak supervision ($e.g.$, image-level labels) for semantic segmentation.
For example, CLIP-ES~\cite{lin2022clip} employs CLIP to refine the class activation map by deigning a softmax function and a class-aware attention-based affinity module for mitigating the category confusion issue. CLIMS~\cite{xie2022clims} employs CLIP knowledge to generate high-quality class activation maps for better weakly-supervised semantic segmentation.

\subsection{Summary and Discussion}

In summary, most VLM studies explore knowledge distillation over two dense visual recognition tasks, namely, object detection and semantic segmenting, where those for the former aim to better align image-level and object-level representations while those for the latter focus on tackling the mismatch between image-level and pixel-level representations. They can also be categorized based on their methodology, including feature-space distillation that enforces embedding consistency between VLM's encoder and the detection (or segmentation) encoder and pseudo-labelling distillation that employs VLM-generated pseudo labels to regularize detection or segmentation models. Moreover, compared with VLM transfer, VLM knowledge distillation has clearly better flexibility of allowing different downstream networks regardless of the original VLMs.

\renewcommand\arraystretch{0.9}
\begin{table*}[ht]
    \centering
    \caption{Performance of VLM pre-training methods over zero-shot prediction setup on image classification tasks. 
    }
    \resizebox{0.95\linewidth}{!}{
    \begin{tabular}{l|c|c|c|c|c|c|c|c|c|c|c|c|c|c}
    \toprule
     Methods & Image encoder & Text encoder & Data Size & \rotatebox{90}{ImageNet-1k~\cite{deng2009imagenet}} & \rotatebox{90}{CIFAR-10~\cite{krizhevsky2009learning}} & \rotatebox{90}{CIFAR-100~\cite{krizhevsky2009learning}} & \rotatebox{90}{Food101~\cite{bossard2014food}} & \rotatebox{90}{sun397~\cite{xiao2010sun}} & \rotatebox{90}{Cars~\cite{krause2013collecting}} & \rotatebox{90}{Aircraft~\cite{maji2013fine}} & \rotatebox{90}{DTD~\cite{cimpoi2014describing}} & \rotatebox{90}{Pets ~\cite{parkhi2012cats}} & \rotatebox{90}{caltech101~\cite{fei2004learning}} & \rotatebox{90}{flowers102~\cite{nilsback2008automated}}\\ 
    \midrule
        CLIP~\cite{radford2021learning} & ViT-L/14 & Transformer & 400M & 76.2 & 95.7 & 77.5 & 93.8 & 68.4 & 78.8 & 37.2 & 55.7 & 93.5 & 92.8 & 78.3\\
        ALIGN~\cite{jia2021scaling}  & EfficientNet & BERT & 1.8B & 76.4 & - & - & - & - & - & - & - & - & - & -\\
        OTTER~\cite{wu2021data} & FBNetV3-C & DeCLUTR-Sci & 3M & - & - & - & - & - & - & - & - & - & - & - \\
        DeCLIP~\cite{li2021supervision} & REGNET-Y & BERT & 88M & 73.7 & - & - & - & - & - & - & - & - & - & - \\
        ZeroVL~\cite{cui2022contrastive} & ViT-B/16 & BERT & 100M & - & - & - & - & - & - & - & - & - & - & - \\
        FILIP~\cite{yao2021filip} & ViT-L/14 & Transformer & 340M & 77.1 & 95.7& 75.3 & 92.2 & 73.1 & 70.8 & 60.2 & 60.7 & 92.0 & 93.0 & 90.1\\
        UniCL~\cite{yang2022unified} & Swin-tiny & Transformer & 16.3M & 71.3 & - & - & - & - & - & - & - & - & - & -\\
        Florence~\cite{yuan2021florence} & CoSwin & RoBERT  & 900M & 83.7& 94.6& 77.6 & 95.1 & 77.0 & 93.2 & 55.5 & 66.4 & 95.9 & 94.7 & 86.2\\
        SLIP~\cite{mu2022slip} & ViT-L & Transformer & 15M & 47.9 & 87.5& 54.2& 69.2& 56.0& 9.0& 9.5& 29.9& 41.6& 80.9& 60.2\\
        PyramidCLIP~\cite{gao2022pyramidclip} & ResNet50 & T5 & 143M & 47.8 & 81.5 & 53.7 & 67.8 & 65.8 & 65.0 & 12.6& 47.2 & 83.7& 81.7& 65.8\\
        Chinese CLIP~\cite{yang2022chinese} & ViT-L/14 & CNRoberta & 200M & -& 96.0& 79.7& - & - & - & 26.2 & 51.2 & -& - & - \\
        LiT~\cite{zhai2022lit} & ViT-g/14 & - & 4B & 85.2 & - & - & - & - & - & - & - & - & - & -  \\
        AltCLIP~\cite{chen2022altclip} & ViT-L/14 & Transformer & 2M & 74.5 & - & - & - & - & - & - & - & - & - & - \\
        FLAVA~\cite{singh2022flava} & ViT-B/16 & ViT-B/16 & 70M & - & - & - & - & - & - & - & - & - & - & - \\
        KELIP~\cite{ko2022large} & ViT-B/32 & Transformer & 1.1B & 62.6& 91.5& 68.6& 79.5& - & 75.4 & - & 51.2 & -& -& -\\
        COCA~\cite{yu2022coca} &  ViT-G/14 & - & 4.8B & 86.3 & - & - & - & - & - & - & - & - & - & -\\
        nCLIP~\cite{zhou2022non} & ViTB/16 & Transformer & 35M & 48.8 & 83.4 &54.5 & 65.8& 59.9 &  18.0 & 5.8 & 57.1 & 33.2 & 73.9& 50.0 \\ 
        K-lite~\cite{shen2022k} & CoSwin & RoBERT5 & 813M & 85.8  & - & - & - & - & - & - & - & - & - & - \\
        NLIP~\cite{huang2022nlip} & ViT-B/16 & BART & 26M & 47.4 & 81.9 & 47.5 & 59.2 & 58.7 & 7.8 & 7.5 & 32.9& 39.2& 79.5& 54.0 \\
        UniCLIP~\cite{leeuniclip} & ViT-B/32 & Transformer &30M & 54.2& 87.8& 56.5& 64.6& 61.1& 19.5& 4.7& 36.6& 69.2& 84.0& 8.0 \\
        PaLI~\cite{chen2022pali}  & ViT-e & mT5 & 12B & 85.4 & - & - & - & - & - & - & - & - & - & - \\ 
        CLIPPO~\cite{tschannen2022image} & ViT-L/16 &  ViT-L/16 & 12B & 70.5 & - & - & - & - & - & - & - & - & - & - \\
         {OneR}~\cite{jang2023unifying} & {ViT-L/16}& {ViT-L/16}& {4M}& {27.3}&-& {31.4}&- &-&-&-&-&-&-&-\\
         {RA-CLIP}~\cite{xie2023ra}& {ViT-B/32}& {BERT}& {15M}& {53.5} &  {89.4} & {62.3} & {43.8} & {46.5}&-&-& {25.6}&-& {76.9}&-\\
         {LA-CLIP}~\cite{fan2023improving}&  {ViT-B/32} & {Transformer}&  {400M}& {64.4}& {92.4}&  {73.0}& {79.7}& {64.9}&  {81.9}& {20.8}&  {55.4}& {87.2}&  {91.8}&  {70.3} \\
         {ALIP}~\cite{yang2023alip}& {ViT-B/32}& {Transformer}& {15M}& {40.3}& {83.8}&  {51.9}& {45.4}& {47.8}& {3.4}& {2.7}& {23.2}& {30.7}& {74.1}& {54.8} \\
         {GrowCLIP}~\cite{deng2023growclip}& {ViT-B/16}& {Transformer}& {12M}& {36.1}& {60.7}&  {28.3}&  {42.5}& {45.5} &-&-& {17.3}&-& {71.9}& {23.3}\\
        \bottomrule

    \end{tabular}
    }
    \label{tab. ZS Performance of existing methods}
\end{table*}

\renewcommand\arraystretch{0.95}
\begin{table}[ht]
    \centering
    \caption{Performance of VLM pre-training methods over zero-shot prediction setup on segmentation tasks.}
    \resizebox{0.96\linewidth}{!}{
    \begin{tabular}{l|c|c|c|c|c|c}
    \toprule
     \multirow{2}{*}{{Method}} & Image & Text & Data & VOC & PASCAL C. & COCO \\ 
      & encoder & encoder & size & \cite{everingham2010pascal} & \cite{mottaghi2014role} &\cite{lin2014microsoft} \\
    \midrule       
        GroupVit~\cite{xu2022groupvit} & ViT & Transformer & 26M & 52.3 & 22.4 & - \\ 
        SegClip~\cite{luo2022segclip} & ViT & Transformer & 3.4M & 52.6 & 24.7& 26.5 \\ 
        \bottomrule

    \end{tabular}
    }
    \label{tab. Performance of existing segmentation methods}
\end{table}

\renewcommand\arraystretch{0.95}
\begin{table}[ht]
    \centering
    \caption{Performance of VLM pre-training methods over zero-shot prediction setup on detection tasks.}
    \resizebox{0.96\linewidth}{!}{
    \begin{tabular}{l|c|c|c|c|c|c}
    \toprule
     \multirow{2}{*}{{Method}} & Image & Text & Data & COCO & LVIS & LVIS Mini. \\ 
      & encoder & encoder & size & \cite{lin2014microsoft} & \cite{gupta2019lvis} & \cite{gupta2019lvis} \\
    \midrule
        RegionClip~\cite{zhong2022regionclip} & ResNet50x4 &Transformer &118k &29.6 &11.3 & -\\ 
        GLIP~\cite{li2022grounded} & Swin-L &BERT  &27.43M &49.8 &26.9 &34.3 \\ 
        FIBER~\cite{doucoarse} &  Swin-B &RoBERTa  &4M &49.3 &-	&32.2  \\
        DetCLIP~\cite{yaodetclip} & Swin-L &BERT &2.43M	&- &35.9 &-  \\      
        \bottomrule

    \end{tabular}
    }
    \label{tab. Performance of existing detection methods}
\end{table}

\begin{figure}[t]
    \centering
    \includegraphics[width=0.42\textwidth]{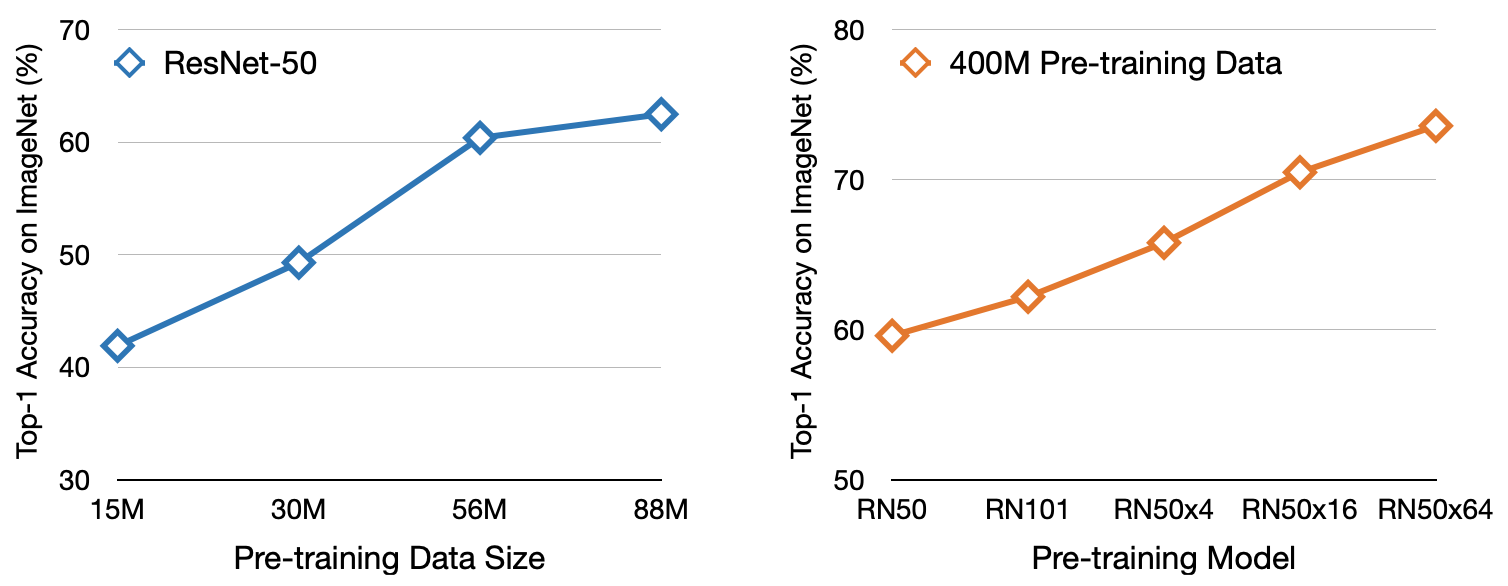}
    \caption{Performance versus data size and model size. It shows that scaling up either the pre-training data~\cite{li2021supervision} or the pre-training model~\cite{radford2021learning} benefits VLM consistently.}
    \label{fig:perf}
\end{figure}

\section{Performance Comparison}\label{sec.benchmark}

In this section, we compare, analyze and discuss the VLM pre-training, VLM transfer learning, and VLM knowledge distillation methods as reviewed in Sections~\ref{Sec.VLP}-\ref{Sec.KD}.

\subsection{Performance of VLM Pre-training}\label{}

As discussed in Sec.~\ref{sec.eval}, \textit{zero-shot prediction} as one widely-adopted evaluation setup assesses VLM generalization over unseen tasks without task-specific fine-tuning. This subsection presents the performance of \textit{zero-shot prediction} over different visual recognition tasks including image classification, object detection, and semantic segmentation.

Table ~\ref{tab. ZS Performance of existing methods} shows evaluations on 11 widely adopted image classification tasks.
Note it shows the best VLM performance as VLM pre-training often have different implementations. Three conclusions can be drawn from Table~\ref{tab. ZS Performance of existing methods} as well as Fig.~\ref{fig:perf}: 1) VLM performance is usually up to the size of training data. As shown in the first graph in Fig.~\ref{fig:perf}, scaling up the pre-training data leads to consistent improvements; 
2) VLM performance is usually up to the model size. As shown in the second graph, with the same pre-training data, 
scaling up model sizes improves the VLM performance consistently; 3) With large-scale image-text training data, VLMs can achieve superior zero-shot performance on various downstream tasks. As Table~\ref{tab. ZS Performance of existing methods} shows, COCA~\cite{yu2022coca} achieves state-of-the-art performance on ImageNet, and FILIP~\cite{yao2021filip} performs well consistently across $11$ tasks.

The superior generalization of VLMs is largely attributed to three factors: 1) Big data - as image-text pairs are almost infinitely available on the Internet, VLMs are usually trained with millions or billions of image and text samples that cover very broad visual and language concepts, leading to strong generalization capability; 2) Big model - compared with traditional visual recognition models, VLMs generally adopt much larger models ($e.g.$, ViT-G in COCA~\cite{yu2022coca} with 2B parameters) that provide great capacity for effective learning from big data; 3) Task-agnostic learning - the supervision in VLM pre-training is usually general and task-agnostic. Compared with task-specific labels in traditional visual recognition, the texts in image-text pairs provide task-agnostic, diverse and informative language supervision which help train generalizable models that works well across various downstream tasks. 

Note several studies~\cite{xu2022groupvit,luo2022segclip,zhong2022regionclip,li2022grounded,doucoarse,yaodetclip} investigate VLM pre-training for object detection and semantic segmentation with local VLM pre-training objectives such as region-word matching~\cite{li2022grounded}. Tables~\ref{tab. Performance of existing segmentation methods} and~\ref{tab. Performance of existing detection methods} summarize \textit{zero-shot prediction} performance on object detection and semantic segmentation tasks. We can observe that VLMs enable effective zero-shot prediction on both dense prediction tasks. Note the results in Tables~\ref{tab. Performance of existing segmentation methods} and~\ref{tab. Performance of existing detection methods} may not be aligned with the conclusions in previous paragraphs, largely because this field of research is under-explored with very limited VLMs on dense visual tasks.

\textbf{Limitations of VLMs.}
As discussed above, although VLMs benefit clearly while data/model size scales up, they still suffer from several limitations: (1) When data/model size keeps increasing, the performance saturates and further scaling up won’t improve performance~\cite{cherti2023reproducible, li2021supervision}; (2) Adopting large-scale data in VLM pre-training necessitates extensive computation resources, $e.g.$, 256 V100 GPUs, 288 training hours in CLIP ViT-L~\cite{radford2021learning}; (3) Adopting large models introduces excessive computation and memory overheads in both training and inference.

\renewcommand\arraystretch{0.9}
\begin{table*}[!t]
    \centering
    \caption{Performance of VLM transfer Learning methods on image classification tasks.}
    \resizebox{0.9\linewidth}{!}{
    \begin{tabular}{l|c|c|c|c|c|c|c|c|c|c|c|c|c|c}
    \toprule
     Methods & Image encoder & Setup &\rotatebox{90}{Average}& \rotatebox{90}{ImageNet-1k~\cite{deng2009imagenet}} & \rotatebox{90}{caltech101~\cite{fei2004learning}} & \rotatebox{90}{Pets~\cite{parkhi2012cats}} & \rotatebox{90}{Cars~\cite{krause2013collecting}} & \rotatebox{90}{Flowers102~\cite{nilsback2008automated}} & \rotatebox{90}{Food101~\cite{bossard2014food}} & \rotatebox{90}{Aircraft~\cite{maji2013fine}} & \rotatebox{90}{SUN397~\cite{xiao2010sun}} & \rotatebox{90}{DTD~\cite{cimpoi2014describing}} & \rotatebox{90}{EuroSAT~\cite{helber2019eurosat}} & \rotatebox{90}{UCF101~\cite{soomro2012ucf101}}\\
    \midrule
        Baseline~\cite{huang2022unsupervised} & ResNet-50 &w/o Transfer& 59.2&60.3&86.1&85.8&55.6&66.1&77.3&16.9&60.2&41.6&38.2&62.7\\
        Baseline~\cite{radford2021learning} & ViT-B/16 & w/o Transfer & 71.7&70.2&95.4&94.1& 68.6&74.8&90.6&31.1&72.2&56.4&60.6&73.5\\
        Baseline~\cite{radford2021learning} & ViT-L/14& w/o Transfer & 73.7 &76.2 & 92.8 &93.5&78.8&78.3&93.8&37.2&68.4&55.7&59.6&76.9\\ 
        \midrule
        CoOp~\cite{zhou2022learning} &ViT-B/16&Few-shot Sup.&71.6& 71.9& 93.7 & 94.5& 68.1& 74.1& 85.2& 28.7& 72.5 & 54.2& 68.7& 67.5 \\ 
        CoCoOp~\cite{zhou2022conditional} &ViT-B/16&Few-shot Sup.&75.8&73.1 &95.8 &96.4 &72.0 &81.7 &91.0 &27.7 &78.3 &64.8 &71.2 &77.6\\ 
        SubPT~\cite{ma2022understanding} &ResNet50&Few-shot Sup.&66.4&63.4&91.7 &91.8 &60.7 &73.8 &81.0 &20.3 &70.2 &54.7 &54.5 &68.1 \\
        LASP~\cite{bulat2022language} &ViT-B/16&Few-shot Sup.&76.1 &73.0 &95.8 &95.7 &72.2 &81.6 &90.5 &31.6 &77.8 &62.8 &74.6 &76.8  \\
        ProDA~\cite{lu2022prompt} &ResNet50&Few-shot Sup.& - &65.3 &91.3 & 90.0&75.5 &95.5 &82.4 &36.6 & - &70.1 &84.3 & -  \\ 
        VPT~\cite{derakhshani2022variational} &ViT-B/16&Few-shot Sup.&77.4 &73.4 &96.4 &96.8 &73.1 &81.1 &91.6 &34.7 &78.5 &67.3 &77.7 &79.0  \\
        ProGrad~\cite{zhu2022prompt} &ResNet-50&Few-shot Sup.&67.9 &62.1 &91.5 &93.4 &62.7 &78.7 &81.0 &21.9 &70.3 &57.8 &59.0 &68.5   \\
        CPL~\cite{he2022cpl} &ViT-B/16&Few-shot Sup.&- &76.0 &96.3 &97.7 &77.2 &81.7 &93.2 & - &80.6 & - & - & -   \\
        PLOT~\cite{chen2022prompt} &ResNet-50&Few-shot Sup.&73.9&63.0 &92.2 &87.2&72.8 &94.8 &77.1 &34.5 &70.0 &65.6 &82.2 &77.3 \\
        CuPL~\cite{pratt2022does} &ViT-L/14 &Few-shot Sup.& - &76.6&93.4&93.8&77.6&-&93.3&36.1&61.7&-&-&-  \\

        UPL~\cite{huang2022unsupervised} &ResNet-50&Unsupervised&68.4 &61.1 &91.4 &89.5 &71.0 &76.6 &77.9 &21.7 &66.4 &55.1 &71.0 &70.2  \\
        TPT~\cite{shutest} &ViT-B/16&Unsupervised&64.8& 69.0&94.2 &87.8 &66.9 &69.0 &84.7 &24.8 &65.5 &47.8 &42.4 &60.8  \\ 
        VP~\cite{bahng2022exploring} &ViT-B/32&Few-shot Sup.& -&- &- &85.0 &- &70.3 &78.9 &- &60.6 &57.1 &96.4 &66.1  \\
        UPT~\cite{zang2022unified} &ViT-B/16 &Few-shot Sup.&76.2&73.2 &96.1 &96.3 &71.8 &81.0 &91.3 &34.5 &78.7 &65.6 &72.0 &77.2  \\
        MaPLE~\cite{khattak2022maple} &ViT-B/16 &Few-shot Sup.&78.6 &73.5 &96.0 &96.6 &73.5 &82.6 &91.4 &36.5 &79.7 &68.2 &82.4 &80.8  \\
        CAVPT~\cite{xing2022class} &ViT-B/16 &Few-shot Sup.&83.2 &72.5 &96.1 &93.5 &88.2 &97.6 &85.0 &57.9 &74.3 &72.6 &92.1 &85.3  \\
        Tip-Adapter~\cite{zhang2021tip} &ViT-B/16&Few-shot Sup.&-&70.8 &- &- &- &- &- &- &- &- &- &-   \\ 
        SuS-X~\cite{udandarao2022sus} &ResNet-50 &Unsupervised& - &61.8 & - & - & - & - & - & - & - & - &45.6 &50.6  \\
        SgVA-CLIP~\cite{peng2022sgva} &ViT-B/16&Few-shot Sup.& -&73.3 &- &- &- &- &- &- &76.4 &- &- &-  \\ 
        VT-Clip~\cite{zhang2021vt} &ResNet-50 &Few-shot Sup.& - & -& -&93.1 &- &- &- &- &- &65.7 &- &-  \\ 
        CALIP~\cite{guo2022calip} &ResNet-50 &Unsupervised &59.4 &60.6 &87.7 &58.6 &77.4 &66.4 &56.3 &17.7 &86.2 &42.4 &38.9&61.7  \\
        Wise-FT~\cite{wortsman2022robust} &ViT-L/14&Supervised&- &87.1 & - & -&- &- &- &- &- &- &- &-  \\ 
         {KgCoOp}~\cite{yao2023visual}& {ViT-B/16}& {Few-shot Sup.}& {74.4}& {70.1}& {94.6}& {93.2}& {71.9}& {90.6}& {86.5}& {32.4}& {71.7}& {58.3}& {71.0}& {78.4}\\
         {ProTeCt}~\cite{wu2023protect}& {ViT-B/16}& {Few-shot Sup.}& {69.9}&-&-&-&-&-&-&-& {74.5}&-&-&-\\
         {RePrompt}~\cite{rong2023retrieval}& {ViT-B/16}& {Few-shot Sup.}& {83.2}& {74.6}& {96.5}& {93.7}& {85.0}& {97.1}& {87.4}& {50.3}& {77.5}& {73.7}& {92.9}& {86.4}\\
         {TaskRes}~\cite{yu2023task}& {ResNet-50}& {Few-shot Sup.}& {75.7} & {65.7}&  {93.4}&  {87.8}&  {76.8}&  {96.0}&  {77.6}&  {36.3}&  {70.6}&  {67.1}&  {84.0}&  {77.9} \\
         {VCD}~\cite{menon2022visual}& {ViT-B/16}& {Unsupervised}&-& {68.0}&-& {86.9}&-&-& {88.5}&-&-& {45.5}& {48.6}&-\\
        \bottomrule[1pt]
    \end{tabular}
    }
    \label{tab:TL Performance of existing methods}
\end{table*}

\subsection{Performance of VLM Transfer Learning}

This section summarizes the performance of VLM transfer under the setups of supervised transfer, few-shot supervised transfer and unsupervised transfer. Table~\ref{tab:TL Performance of existing methods} shows the results on 11 widely adopted image classification datasets ($e.g.$, EuroSAT~\cite{helber2019eurosat}, UCF101~\cite{soomro2012ucf101}) with different backbones such as CNN backbone ResNet-50 and Transformer backbones ViT-B and ViT-L. Note Table~\ref{tab:TL Performance of existing methods} summarizes the performance of 16-shot setup for all \textit{few-shot supervised} methods.

Three conclusions can be drawn from Table~\ref{tab:TL Performance of existing methods}. First, VLM transfer setups 
helps in downstream tasks consistently. For example, supervised Wise-FT, few-shot supervised CoOp and unsupervised TPT improve accuracy by 10.9\%,1.7\% and 0.8\%, respectively, on ImageNet. As pre-trained VLMs generally suffer from domain gaps with task-specific data, VLM transfer can mitigate the domain gaps by learning from task-specific data, being labelled or unlabelled. 

Second, the performance of few-shot supervised transfer lag far behind that of supervised transfer ($e.g.$, 87.1\% in WiseFT~\cite{wortsman2022robust} and 76.6\% in CuPL~\cite{pratt2022does}), largely because VLMs may overfit to few-shot labelled samples with degraded generalization. Third, unsupervised transfer can perform comparably with few-shot supervised transfer ($e.g.$, unsupervised UPL~\cite{huang2022unsupervised} outperforms 2-shot supervised CoOp~\cite{zhou2022learning} by 0.4\%, unsupervised TPT~\cite{shutest} is comparable with 16-shot CoOp~\cite{zhou2022learning}), largely because unsupervised transfer can access massive unlabelled downstream data with much lower overfitting risks. Nevertheless, unsupervised transfer also faces several challenges such as noisy pseudo labels. We expect more studies on this promising but changeling research direction.

\subsection{Performance of VLM Knowledge Distillation}
This section presents how VLM knowledge distillation helps in the tasks of object detection and semantic segmentation. Tables~\ref{tab:KD OD Performance of existing methods} and~\ref{tab:KD Seg Performance of existing methods} show the knowledge distillation performance on the widely used detection datasets ($e.g.$, COCO~\cite{lin2014microsoft} and LVIS~\cite{gupta2019lvis}) and segmentation datasets ($e.g.$, PASCAL VOC~\cite{everingham2010pascal} and ADE20k\cite{zhou2017scene}), respectively. We can observe that VLM knowledge distillation brings clear performance improvement on detection and segmentation tasks consistently, largely because it introduces general and robust VLM knowledge while benefiting from task-specific designs in detection and segmentation models.

\renewcommand\arraystretch{0.95}
\begin{table}[!t]
    \centering
    \caption{Performance of VLM knowledge distillation on object detection. CLIP Transformer is CLIP text encoder.}
    \resizebox{0.99\linewidth}{!}{
    \begin{tabular}{l|c|c|c|c|c|c|c|c}
    \toprule
    \multirow{2}{*}{Method} & Vision-Language & \multicolumn{3}{c|}{COCO~\cite{lin2014microsoft}} & \multicolumn{4}{c}{LVIS~\cite{gupta2019lvis}} \\
    \cmidrule{3-9}
    &Model&AP$_{base}$&AP$_{novel}$&AP&AP$_{r}$&AP$_{c}$&AP$_{f}$&AP\\
    \midrule
    Baseline~\cite{gu2021open} & -&28.3&26.3&27.8 & 19.5 & 19.7 & 17.0 & 18.6\\
     ViLD~\cite{gu2021open} & CLIP ViT-B/32 & 59.5& 27.6 & 51.3&16.7& 26.5& 34.2& 27.8 \\ 
    DetPro~\cite{du2022learning} & CLIP ViT-B/32 &-&-&34.9 &20.8& 27.8& 32.4& 28.4  \\
    HierKD~\cite{ma2022open} &CLIP ViT-B/32 &53.5 &27.3&- 
 &-&-&-&- \\ 
    RKD~\cite{rasheedbridging} &CLIP ViT-B/32 & 56.6& 36.9& 51.0&21.1& 25.0& 29.1& 25.9 \\ 
    PromptDet~\cite{feng2022promptdet} & CLIP Transformer & -&26.6 & 50.6&21.4& 23.3& 29.3& 25.3\\ 
    PB-OVD~\cite{gao2022open} &CLIP Transformer &46.1& 30.8& 42.1& -&-&-&-\\
    CondHead~\cite{wang2022learning} & CLIP ViT-B/32& 60.8& 29.8& 49.0& 18.8& 28.3& 33.7& 28.8  \\
    VLDet~\cite{lin2022learning} & CLIP Transformer & 50.6& 32.0& 45.8& 26.3& 39.4& 41.9& 38.1\\
    F-VLM~\cite{kuo2022f} & CLIP ResNet-50&-&28.0& 39.6&32.8&-&-& 34.9 \\ 
    OV-DETR~\cite{zang2022open} & CLIP ViT-B/32 &52.7 &29.4 &61.0 & 17.4& 25.0& 32.5& 26.6 \\ 
    Detic~\cite{zhou2022detecting} & CLIP Transformer& 45.0& 27.8& 47.1& 17.8& 26.3& 31.6& 26.8  \\ 
    OWL-ViT~\cite{minderer2022simple} & CLIP ViT-B/32 &-&-&28.1&18.9&-&-&22.1   \\ 
    VL-PLM~\cite{zhao2022exploiting} & CLIP ViT-B/32 & 60.2 &34.4& 53.5&-&-&-&22.2  \\ 
    P$^3$OVD~\cite{long2022p} & CLIP ResNet-50 &51.9& 31.5& 46.6&-&-&-&10.6 \\
     {RO-ViT}~\cite{kim2023region} &  {CLIP ViT-L/16} & -& {33.0}& {47.7}& {32.1}&-&-& {34.0}\\
     {BARON}~\cite{wu2023aligning} & {CLIP ResNet-50} &  {54.9} & {42.7} &  {51.7} & {23.2} & {29.3}&  {32.5}&  {29.5}\\
     {OADP}~\cite{wang2023object}& {CLIP ViT-B/32} &   {53.3}& {30.0}& {47.2}& {21.9} & {28.4}&  {32.0}&  {28.7} \\
    \bottomrule
    \end{tabular}
    }
    \label{tab:KD OD Performance of existing methods}
\end{table}

\renewcommand\arraystretch{0.99}
\begin{table}[!t]
    \centering
    \caption{Performance of VLM knowledge distillation on semantic segmentation tasks.}
    \resizebox{0.96\linewidth}{!}{
    \begin{tabular}{l|c|c|c|c|c|c|c}
    \toprule
    \multirow{2}{*}{{Method}}& Vision-Language& A-847& PC-459& A-150& PC-59& PAS-20& C-19\\
    & Model & \cite{zhou2017scene} & \cite{mottaghi2014role} & \cite{zhou2017scene} &\cite{mottaghi2014role} & \cite{everingham2010pascal}  & \cite{cordts2016cityscapes} \\
    \midrule
    Baseline~\cite{xian2019semantic} & - & - &- &- &24.3 &18.3 &-\\
    LSeg    ~\cite{ding2022decoupling} & CLIP ResNet-101 &- &- &- &- &47.4 &-\\
    ZegFormer~\cite{li2021language}  & CLIP ResNet-50 &- &- &16.4 &- &80.7 &-\\
    OVSeg~\cite{liang2022open} &CLIP Swin-B & 9.0 & 12.4& 29.6& 55.7& 94.5& - \\
    ZSSeg~\cite{xu2022simple} &CLIP ResNet-101 &7.0 &-&20.5&47.7&-&34.5 \\ 
    OpenSeg~\cite{ghiasi2022scaling} & CLIP Eff-B7 &6.3 &9.0 &21.1 &42.1 &-&- \\ 
    ReCo~\cite{shinreco} &CLIP ResNet-101 &- &-&-&-&-&24.2\\     
     {FreeSeg}~\cite{qin2023freeseg}&  {CLIP ViT-B/16} &-&-& {39.8}&-& {86.9}&-\\
    \bottomrule
    \end{tabular}
    }
    \label{tab:KD Seg Performance of existing methods}
\end{table}

\subsection{Summary} 
Several conclusions can be drawn 
from
Tables~\ref{tab. ZS Performance of existing methods}-\ref{tab:KD Seg Performance of existing methods}. Regarding \textit{performance}, VLM pre-training achieves remarkable zero-shot prediction on a wide range of image classification tasks due to its well-designed pre-training objectives. Nevertheless, the development of VLM pre-training for dense visual recognition tasks (on region or pixel-level detection and segmentation) lag far behind. 
In addition, VLM transfer has made remarkable progress across multiple image classification datasets and vision backbones. However, supervised or few-shot supervised transfer still requires labelled images, whereas the more promising but challenging unsupervised VLM transfer has been largely neglected.

Regarding \textit{benchmark}, most VLM transfer studies adopt the same pre-trained VLM as the baseline model and perform evaluations on the same downstream tasks, which facilitates benchmarking greatly. 
They also release their codes and do not require intensive computation resources, easing reproduction and benchmarking greatly. Differently, VLM pre-training has been studied with different data ($e.g.$, CLIP~\cite{radford2021learning}, LAION400M~\cite{schuhmann2021laion} and CC12M~\cite{changpinyo2021conceptual}) and networks ($e.g.$, ResNet~\cite{he2016deep}, ViT~\cite{dosovitskiy2020image}, Transformer~\cite{vaswani2017attention} and BERT~\cite{devlin2018bert}), making fair benchmarking a very challenging task. Several VLM pre-training studies also use non-public training data~\cite{radford2021learning,yao2021filip,chen2022pali} or require intensive computation resources ($e.g.$, 256 V100 GPUs in ~\cite{radford2021learning}). 
For VLM knowledge distillation, many studies adopt different task-specific backbones ($e.g.$, ViLD adopts Faster R-CNN, OV-DETR uses DETR) which complicates benchmarking greatly. Hence, VLM pre-training and VLM knowledge distillation are short of certain norms in term of training data, networks and downstream tasks.

\section{FUTURE DIRECTIONS}\label{sec.future}

VLM enables effective usage of web data, zero-shot prediction without any task-specific fine-tuning, and open-vocabulary visual recognition of images of arbitrary categories. It has been achieving great success with incredible visual recognition performance. In this section, we humbly share several research challenges and potential research directions that could be pursued in the future VLM study on various visual recognition tasks.

For \textbf{VLM pre-training}, there are four challenges and potential research directions as listed.

\textit{(1) Fine-grained vision-language correlation modelling.} With local vision-language correspondence knowledge~\cite{li2022grounded,yaodetclip}, VLMs can better recognize patches and pixels beyond images, greatly benefiting dense prediction tasks such as object detection and semantic segmentation that play an important role in various visual recognition tasks. Given the very limited VLM studies along this direction~\cite{xu2022groupvit, luo2022segclip,zhong2022regionclip,li2022grounded,doucoarse,yaodetclip}, we expect more research in fine-grained VLM pre-training for zero-shot dense prediction tasks.

\textit{(2) Unification of vision and language learning.} The advent of Transformer~\cite{vaswani2017attention,dosovitskiy2020image} makes it possible to unify image and text learning within a single Transformer by tokenizing images and texts in the same manner. 
Instead of employing two separate networks as in existing VLMs~\cite{radford2021learning, jia2021scaling}, unifying vision and language learning enables efficient communications across data modalities which can benefit both training effectiveness and training efficiency. This issue has attracted some attention~\cite{tschannen2022image,jang2023unifying} but more efforts are needed towards more sustainable VLMs.

\textit{(3) Pre-training VLMs with multiple languages.} Most existing VLMs are trained with a single language ($i.e.$, English)~\cite{radford2021learning,jia2021scaling}, which could introduce bias in term of cultures and regions~\cite{sharma2018conceptual, changpinyo2021conceptual} and hinder VLM applications in other language areas. Pre-training VLMs with texts of multiple languages~\cite{ko2022large,chen2022altclip} allows learning different cultural visual characteristics for the same meaning of words but different languages~\cite{schuhmann2022laion}, enabling VLMs to work efficiently and effectively across different language scenarios. We expect more research on multilingual VLMs.

\textit{(4) Data-efficient VLMs.} Most existing work trains VLMs with large-scale training data and intensive computations, making its sustainability a big concern. Training effective VLMs with limited image-text data can mitigate this issue greatly. For example, instead of merely learning from each image-text pair, more useful information could be learned with the supervision among image-text pairs~\cite{wu2021data,li2021supervision}.

\textit{(5) Pre-training VLMs with LLMs.} Recent studies~\cite{fan2023improving,yang2023alip} retrieve rich language knowledge from LLMs to enhance VLM pre-training. Specifically, they employ LLMs to augment the texts in the raw image-text pairs, which provides richer language knowledge and helps better learn vision-language correlation.
We expect more exploration of LLMs in VLM pre-training in the future research.

For \textbf{VLM Transfer Learning}, there are three challenges and potential research directions as listed.

\textit{(1) Unsupervised VLM transfer.} Most existing VLM transfer studies work with a supervised or few-shot supervised setup that requires labelled data, and the latter tends to overfit to the few-shot samples. Unsupervised VLM transfer allows exploring massive unlabelled data with much lower risk of overfitting. More studies on unsupervised VLM transfer are expected in the ensuing VLM studies.

\textit{(2) VLM transfer with visual prompt/adapter.} Most existing studies on VLM transfer focus on text prompt learning~\cite{zhou2022learning}. Visual prompt learning or visual adapter, which is complementary to text prompting and can enable pixel-level adaptation in various dense prediction tasks, is largely neglected. More VLM transfer studies in visual domain are expected.

\textit{(3) Test-time VLM transfer.} Most existing studies conduct transfer by fine-tuning VLMs on each downstream task ($i.e.$, prompt learning), leading to repetitive efforts while facing many downstream tasks. Test-time VLM transfer allows adapting prompts on the fly during inference, circumventing the repetitive training in existing VLM transfer. We can foresee more studies on test-time VLM transfer.

\textit{(4) VLM transfer with LLMs.}  
Different from prompt engineering and prompt learning, several attempts~\cite{pratt2022does,menon2022visual} exploit LLMs~\cite{brown2020language} to generate text prompts that better describe downstream tasks. This approach is automatic and requires little labelled data. 
We expect more exploration of LLMs in VLM transfer in the future research.

\textbf{VLM knowledge distillation} could be further explored from two aspects. The first is knowledge distillation from multiple VLMs that could harvest their synergistic effect by coordinating knowledge distillation from multiple VLMs. The second is knowledge distillation for other visual recognition tasks such as instance segmentation, panoptic segmentation, person re-identification etc. 

\section{Conclusion}

Vision-language models for visual recognition enables effective usage of web data and allows zero-shot predictions without task-specific fine-tuning, which is simple to implement yet has achieved great success on a wide range of recognition tasks. This survey extensively reviews vision-language models for visual recognition from several perspectives, including background, foundations, datasets, technical approaches, benchmarking, and future research directions.
The comparative summary of the VLM datasets, approaches, and performance in tabular forms provides a clear big picture of the recent development in VLM pre-training which will greatly benefit the future research along this emerging but very promising research direction.

\ifCLASSOPTIONcaptionsoff
  \newpage
\fi



%

{\small
\bibliographystyle{revision_ref}
\bibliography{ref}
}

\appendix

\section*{A. Statistics on Visual Recognition VLM Publications}

As shown in Figure 1 (in the main manuscript), we count the number of visual recognition VLM
publications on Google Scholar over the recent two years.
Specifically, we consider all the papers that have cited the pioneer VLM study ($i.e.$, CLIP) as potential publications and identify a publication as the visual recognition VLM study if it contains any one of keywords image classification, object detection and semantic segmentation.
For the year 2023, we project the total publications based on the number of publications from 1 Jan 2023 to 30 November 2023.

\section*{B. Datasets for Pre-training VLM}

For VLM pre-training, multiple large-scale image-text datasets~\cite{radford2021learning,jia2021scaling,schuhmann2021laion,schuhmann2022laion} were collected from  
the internet.
Compared with traditional crowd-labelled datasets~\cite{deng2009imagenet,cordts2016cityscapes,everingham2010pascal}, the image-text datasets~\cite{schuhmann2021laion,radford2021learning} 
are much larger and cheaper to collect. 
For example, recent image-text datasets are generally at billion scale~\cite{schuhmann2021laion,schuhmann2022laion,chen2022pali}.
Beyond image-text datasets, several studies~\cite{li2022grounded,yu2022coca,yaodetclip,tschannen2022image} utilize auxiliary datasets to provide additional information for better vision-language 
modelling, $e.g.$, GLIP~\cite{li2022grounded} leverages Object365~\cite{shao2019objects365} for extracting region-level features.

\subsection*{B.1. Image-Text Datasets}
\begin{itemize}

    \item \textbf{SBU~\cite{ordonez2011im2text}} contains 1M images collected from Flicker website, paired with visually relevant captions.
    
    \item \textbf{COCO Caption~\cite{chen2015microsoft}} contains over 330k images from MS COCO~\cite{lin2014microsoft}. It has two versions: COCO Caption c5 with 5 reference captions for 330k images and COCO Caption c40 that provides 40 reference captions for a randomly sampled subset of 5,000 images. 
    
    \item \textbf{YFCC100M~\cite{thomee2016yfcc100m}} is a multimedia dataset containing 99.2M images and 0.8M videos with texts.

    \item \textbf{VG~\cite{krishna2017visual}} provides a multi-perspective understanding of images, $e.g.$, object-level information, scene graphs and visual question answer pairs. VG contains 108,000 images, each with 50 descriptions.

    \item \textbf{CC3M~\cite{sharma2018conceptual}} is an image captioning dataset which consists of about 3.3M image-text pairs from the web.

    \item \textbf{CC12M~\cite{changpinyo2021conceptual}} is introduced specifically for VLM pre-training. By relaxing the data collection pipeline used in CC3M~\cite{sharma2018conceptual}, CC12M collects less precise but much larger size of data, $i.e.$, 12M image-text pairs.
    
    \item \textbf{LR~\cite{pont2020connecting}} is an image captioning dataset with local multi-modal annotations, where every word is localized in the image with a mouse trace segment. It contains 848,749 images with 873,107 captions.
    
    \item \textbf{WIT~\cite{srinivasan2021wit}} is a large multi-modal multilingual dataset collected from Wikipedia, which consists of 37.6M image-text pairs across 108 languages.

    \item \textbf{Red Caps~\cite{desai2021redcaps}} is a image-text dataset collected from social media Reddit, which contains 12M image-text pairs covering various objects and scenes.

    \item \textbf{LAION400M~\cite{schuhmann2021laion}:} LAION400M consists of 400M image-text pairs filtered by CLIP~\cite{radford2021learning}, which also provides the data embeddings and kNN indices.
    
    \item \textbf{LAION5B~\cite{schuhmann2022laion}} contains over 5.8B image-text pairs, which consists of three parts: 2.32B English image-text pairs, 2.26B multilingual image-text pairs and 1.27B pairs without specific language.

    \item \textbf{WuKong~\cite{gu2022wukong}} is a large-scale Chinese multi-modal dataset, which contains 100M Chinese image-text pairs collected from the web.

    \item \textbf{CLIP~\cite{radford2021learning}} is a large-scale web image-text dataset, which contains 400M image-text pairs collected form a variety of publicly available sources on the internet.
    
    \item \textbf{ALIGN~\cite{jia2021scaling}} is an image-text dataset, which contains 1.8B noisy image-text pairs covering board concepts.

    \item \textbf{FILIP~\cite{yao2021filip}} is a large-scale image-text dataset with 300M image-text pairs collected from the internet.

    \item \textbf{WebLI~\cite{chen2022pali}} is a multilingual image-text dataset collected from the web, which comprises 10B images with 12B corresponding texts across 109 languages.
\end{itemize}

\subsection*{B.2. Auxiliary Datasets}

\begin{itemize}
    \item \textbf{JFT3B~\cite{zhai2022scaling}} contains nearly 3B images annotated with a noisy class hierarchy of around 30k labels.

    \item \textbf{C4~\cite{raffel2020exploring}} is a collection of about 750GB English text sourced from the public Common Crawl web scrape.

    \item \textbf{Object365~\cite{shao2019objects365}} is a object detection dataset with 365 categories, 638K images, and $\sim$10M bounding boxes.

    \item \textbf{Gold-G~\cite{kamath2021mdetr}} is an object-phrase dataset for object detection, which includes 0.8M human-annotated visual grounding data curated by~\cite{kamath2021mdetr}.

\end{itemize}

\section*{C. Datasets for Evaluation}

Many visual recognition datasets have been adopted for VLM evaluations as shown in Table 2 (in the main manuscript) including $27$ image classification datasets, $4$ object detection datasets, $4$ semantic segmentation datasets, $2$ image-text retrieval datasets and $3$ action recognition datasets.
Below please find the detail of each dataset.

\subsection*{C.1. Datasets for Image Classification}

\begin{itemize}
    \item \textbf{Food-101~\cite{bossard2014food}} is a real-world food dataset for fine-grained recognition. The dataset consists of 101,000 images, covering 101 classes.  
    Specifically, every class contains 250 cleaned test samples and 750 purposely uncleaned training samples. 
    
    \item \textbf{CIFAR-10~\cite{krizhevsky2009learning}} contains a set of small images, which is commonly used for the image classification tasks. This dataset includes 60000 images with size 32 by 32, annotated with ten categories. This dataset has been divided into 5000 training samples and 1000 testing samples per class. 
    
     \item \textbf{CIFAR-100~\cite{krizhevsky2009learning}} is almost the same as CIFAR10, except CIFAR-100 instead contains 60000 samples with 100 categories that have been grouped into twenty super-categories.

    \item \textbf{Birdsnap~\cite{berg2014birdsnap}} is a fine-grained classification dataset collected from Flicker. There are 49,829 images belonging to 500 bird species, including 47,386 training images and 2433 testing images. In this dataset, every image has been labelled with a bounding box, the coordinates of 17 parts, and auxiliary attribute annotations like male, female, immature, etc.

    \item \textbf{SUN397~\cite{xiao2010sun}} is proposed for scene recognition and contains 39700 images covering 397 well-sampled categories. The scene classification performance by human is provided as the reference for the comparisons with computational methods.

    \item \textbf{Stanford Cars~\cite{krause2013collecting}} is designed for fine-grained recognition, containing 16185 images covering 196 categories. This dataset has been divided into 8,144 training samples and 8,041 testing samples.

    \item \textbf{FGVC Aircraft~\cite{maji2013fine}} includes 10K samples spanning 100 aircraft model variants. Every samples is labeled with a tight bounding box and a hierarchical category annotation. This dataset has been equally separated into training, validation, and test subsets, where every subset contains 33 or 34 images per variant. 

    \item \textbf{PASCAL VOC 2007 Classification~\cite{everingham2010pascal}} is the widely-used dataset for various visual recognition tasks like detection, segmentation, and classification. There are 9963 samples covering 20 classes, including 5011 training images and 4952 testing images. Every sample in PASCAL VOC 2007 contains pixel-wise labels, object-level labels with object box and category labels.

    \item \textbf{Describable Textures~\cite{cimpoi2014describing} (DTD)} is a collection of textural images for image recognition. This dataset includes 5640 samples with forty seven categories, which has been equally separated into training, validation, and test subsets, where each subset contains 40 images per class. For each image, the main category and a list of the joint attributes are provided.

    \item \textbf{Oxford-IIIT PETS~\cite{parkhi2012cats}} includes 7,349 cat and dog images with thirty seven different breeds, in which twenty five are dog breeds and twelve are cat breeds. These samples are separated into the training subset with around 1850 samples, the validation subset with about 1850 samples and testing subset with approximate 3700 samples.
    Every sample has been annotated with a breed annotation, a pixel-wise annotation that marks the body, and a rectangle box for locating the head.

    \item \textbf{Caltech-101~\cite{fei2004learning}} consists of 9145 images belonging to 101 classes, Every category includes 40-80 images. For each image, the dataset provides an annotation to segment the foreground object.

    \item \textbf{Oxford 102 Folwers~\cite{nilsback2008automated}} is proposed for fine-grained image classification. This dataset contains 8189 flowers images that belong to 102 species. Each category contains 40-200 samples, including the flower captured under various sizes and illumination environments. Besides, this dataset also contains pixel-wise annotations.
    
   \item \textbf{Facial Emotion Recognition 2013~\cite{goodfellow2013challenges}} is collected by requesting images associated with 184 key emotional terms from Google. The dataset contains 35,887 grayscale images with a resolution of 48x48 pixels and with 7 types of emotions.

    \item \textbf{STL-10~\cite{coates2011analysis}} is a type of classification benchmark for researching on unsupervised and self-taught training. It includes 10 categories and an unsupervised training subset with 100K samples, a supervised training subset with 5K samples, and a testing subset with 8K samples.

    \item \textbf{EuroSAT~\cite{helber2019eurosat}} is a set of satellite images used to benchmark the land use and land cover recognition tasks. It covers thirteen frequency bands with ten categories of 27K annotated and geo-referenced samples. Two datasets are provided including the RGB image dataset and the multi-spectral image dataset.

    \item \textbf{RESISC45~\cite{cheng2017remote}} has been proposed to benchmark Remote Sensing Image Scene Classification (RESISC). This dataset includes 31,500 sample with the image size of 256 by 256 and forty five scene categories, every category containing 700 samples. 
    Besides, RESISC45 covers a wide range of spatial resolutions from 20cm to over 30m per pixel.
    
    \item \textbf{GTSRB~\cite{stallkamp2011german}} is a dataset for traffic signs classification, containing 50,000 images taken from various street scenes in Germany. It is classified into 43 categories, including a training subset with 39,209 samples and a testing subset with 12,630 samples.

    \item \textbf{Country211~\cite{radford2021learning}} is an image classification dataset for geolocation evaluation, which is a subset of the YFCC100M dataset. For each country, there are one hundred and fifty train samples, fifty validation samples, and one hundred test samples.

    \item \textbf{PatchCamelyon~\cite{veeling2018rotation}} 
    includes 327,680 RGB images with the size of 96 by 96 from Camelyon16, with a training subset with 75\% samples, a validation subset with 12.5\% samples, and a testing subset with 12.5\% samples. 
    Every sample has been labelled with a binary annotation showing if it contains the metastatic tissue.

    \item \textbf{Hateful Memes~\cite{kiela2020hateful}} has been proposed for hateful meme classification ($i.e.$, image with text) created by Facebook AI. It includes over 10k memes annotated with either the hateful label or the non-hateful label.

    \item \textbf{Rendered SST2~\cite{radford2021learning}} has been proposed for benchmarking optical character recognition. It includes 2 categories (the categoryies of positives and negatives). This dataset has been separated into 3 subsets: a training subset with 6920 samples, a validation subset with 872 samples, and a test subset with 1821 samples.

    \item \textbf{ImageNet-1k~\cite{deng2009imagenet}} includes about 1.2M samples that are uniformly distributed across the one thousand categories. The category annotation of ImageNet-1k follows WordNet hierarchy and every sample is annotated with one category label. 
    Bisides, ImageNet-1k is one of the most popular image classification benchmarks.

    \item \textbf{CLEVR Counts~\cite{johnson2017clevr}} is a subset of CLEVR dataset, which is designed for visual question answering to evaluate the ability to perform visual reasoning. The counting tasks include 2000 training samples and 500 test samples. 

    \item \textbf{SVHN~\cite{netzer2011reading}} is a dataset for recognizing digits and numbers in real-world images which are collected from Google Street View images. It consists of about 600,000 images and all digits are cropped from the images and resized to a fixed resolution of 32x32 pixels.

    \item \textbf{IIIT5k~\cite{mishra2012scene}} contains 5,000 cropped word images collected from Google image search by using search keyword such as signboards, house name plates, and movie posters, etc.
    The dataset is split into two parts, $i.e.$, 2,000 word images for training and 3,000 word images for validation, respectively.

    \item \textbf{Rendered SST2~\cite{socher2013recursive}} is sentiment classification dataset which consists of two sentiment categories, $i.e.$, negative and positive.
    The sentences in the dataset are extracted from Stanford Sentiment Treebank dataset.

\end{itemize}

\subsection*{C.2. Datasets for Action Recognition}

\begin{itemize}

    \item \textbf{UCF101~\cite{soomro2012ucf101}} has been proposed for benchmarking human action recognition with videos.    
    It includes about 13k video clips of 101 actions, which are collected from YouTube. 
    The video clips in the dataset have a resolution of 320x240 pixels and a frame rate of 25 FPS.

    \item \textbf{Kinetics700~\cite{carreira2019short}} is a video dataset for recognizing human action. 
    It consists of about 65,000 video clips with 700 human actions, where each action category has more than 700 video clips lasting around 10 seconds.

    \item \textbf{RareAct\cite{miech2020rareact}} is a video dataset designed for identifying rare actions such as ``Unplug Oven'' and ``fry phone''. This dataset aims to evaluate action recognition models on unlikely combinations of common action verbs and object nouns.
    It contains 122 human actions, where the verbs and object nouns in actions are rarely co-occurring together in HowTo100M.

\end{itemize}

\subsection*{C.3. Datasets for Semantic Segmentation}

\begin{itemize}

    \item \textbf{ADE20k~\cite{zhou2017scene}} is a semantic segmentation dataset which consists of 150 classes. It consists of a training subset with 25,574 samples and a validation subset with 2,000 samples.

    \item \textbf{PASCAL VOC 2012 Segmentation~\cite{everingham2010pascal}} contains 20 categories including vehicles, household and animals. This dataset includes a training subset with 1,464 samples and a testing subset with 1,449 samples, all of which are with pixel-wise annotations.

    \item \textbf{PASCAL Content~\cite{mottaghi2014role}:} PASCAL Content is an extension of PASCAL VOC 2010 detection dataset~\cite{everingham2010pascal}, which contains more than 400 categories with pixel-wise annotations. It has 4,998 training images and 1,449 validation images.

    \item \textbf{Cityscapes~\cite{cordts2016cityscapes}:} Cityscapes is a dataset for the visual recognition on street scenes. This dataset includes a training subset with 2,975 samples and a testing subset with 500 samples, all of which are with pixel-wise annotations of 19 categories.

\end{itemize}

\subsection*{C.4. Datasets for Object Detection}

\begin{itemize}

    \item \textbf{MS COCO~\cite{lin2014microsoft}:} MS COCO Dataset is a dataset for object detection. It consists of two versions: MS COCO 2014 contains 83,000 training images and 41,000 validation images with bounding box annotations of 80 categories and MS COCO 2017 contains 118,000 training images and 5,000 validation images with bounding box annotations of 80 categories.

    \item \textbf{ODinW~\cite{li2022elevater}:} ODinW is a benchmark to evaluate the task-level transfer ability of pre-trained vision models, which consist of 35 free public Object Detection datasets in various domains. The dataset contains 132k training images and 20K testing images belonging to 314 concepts. Also, many of the 35 tasks have very limited (less than 100) training images, which makes it extremely difficult for standard detectors without any pre-training.
    
    \item \textbf{LVIS~\cite{gupta2019lvis}:} LVIS is a large vocabulary dataset for long-tailed instance detection/ segmentation. The dataset contains 1203 categories with federated human annotations on 100K images.

\end{itemize}

\subsection*{C.5. Datasets for Image and Text Retrieval}

\begin{itemize}

    \item \textbf{Flickr30k~\cite{young2014image}:} Flickr30K is a dataset for automatic image description and grounded language understanding. It contains 31,000 images collected from Flickr, where each image is provided with 5 captions.

    \item \textbf{COCO Caption~\cite{chen2015microsoft}:} COCO Caption contains over 330k images from MS COCO~\cite{lin2014microsoft}. It has two versions: COCO Caption c5 with 5 reference captions for 330k images and COCO Caption c40 that provides 40 reference captions for a randomly sampled subset of 5,000 images.

\end{itemize}
%

\vspace{-1cm}

\begin{IEEEbiographynophoto}{Jingyi Zhang} received her B.Sc. degree in electronic information science and technology from the University of Electronic Science and Technology of China (UESTC) and M.Sc. degree in signal processing from the Nanyang Technological University (NTU). She is currently a Research Associate and Ph.D. student with School of Computer Science and Engineering, NTU. Her research interests include computer vision, object detection.
\end{IEEEbiographynophoto}
\vspace{-1cm}

\begin{IEEEbiographynophoto}{Jiaxing Huang} received his B.Eng. and M.Sc. in EEE from the University of Glasgow, UK, and the Nanyang Technological University (NTU), Singapore, respectively. He is currently a Research Associate and Ph.D. student with School of Computer Science and Engineering, NTU, Singapore. His research interests include computer vision and machine learning.
\end{IEEEbiographynophoto}
\vspace{-1cm}

\begin{IEEEbiographynophoto}{Sheng Jin} is currently a Research Fellow at Nanyang Technology University (NTU), Singapore. Before that, he received his B.Sc. degree in Applied Mathematics from Harbin Institute of Technology and the Ph.D. degree in Computer Science and Technology om Harbin Institute of Technology. His research interests include computer vision and machine learning.
\end{IEEEbiographynophoto}
\vspace{-1cm}

\begin{IEEEbiographynophoto}{Shijian Lu} is an Associate Professor with the School of Computer Science and Engineering at the Nanyang Technological University, Singapore. He received his PhD in electrical and computer engineering from the National University of Singapore. His major research interests include image and video analytics, visual intelligence, and machine learning.
\end{IEEEbiographynophoto}

\end{document}